\documentclass[10pt,twocolumn,letterpaper]{article}

\usepackage{iccv}
\usepackage{times}
\usepackage{epsfig}
\usepackage{graphicx}
\usepackage{amsmath}
\usepackage{amssymb}
\usepackage{subcaption}
\usepackage{diagbox}

\usepackage[pagebackref=true,breaklinks=true,letterpaper=true,colorlinks,bookmarks=false]{hyperref}

\iccvfinalcopy % *** Uncomment this line for the final submission

\begin{document}

%%%%%%%%% TITLE
\title{End-to-End Monocular Vanishing Point Detection Exploiting Lane Annotations}

\author{Hiroto Honda, Motoki Kimura, Takumi Karasawa, Yusuke Uchida\\
Mobility Technologies\\
{\tt\small \{hiroto.honda, motoki.kimura, takumi.karasawa, yusuke.uchida\}@mo-t.com}
}

\maketitle
\ificcvfinal\thispagestyle{empty}\fi

%%%%%%%%% ABSTRACT
\begin{abstract}
   Vanishing points (VPs) play a vital role in various computer vision tasks, especially for recognizing the 3D scenes from an image. In the real-world scenario of automobile applications, it is costly to manually obtain the external camera parameters when the camera is attached to the vehicle or the attachment is accidentally perturbed. 
   In this paper we introduce a simple but effective end-to-end vanishing point detection. By automatically calculating intersection of the extrapolated lane marker annotations, we obtain geometrically consistent VP labels and mitigate human annotation errors caused by manual VP labeling. With the calculated VP labels we train end-to-end VP Detector via heatmap estimation. The VP Detector realizes higher accuracy than the methods utilizing manual annotation or lane detection, paving the way for accurate online camera calibration. 
\end{abstract}

%%%%%%%%% BODY TEXT
\begin{figure}[t]
\begin{center}
   \includegraphics[width=0.99\linewidth]{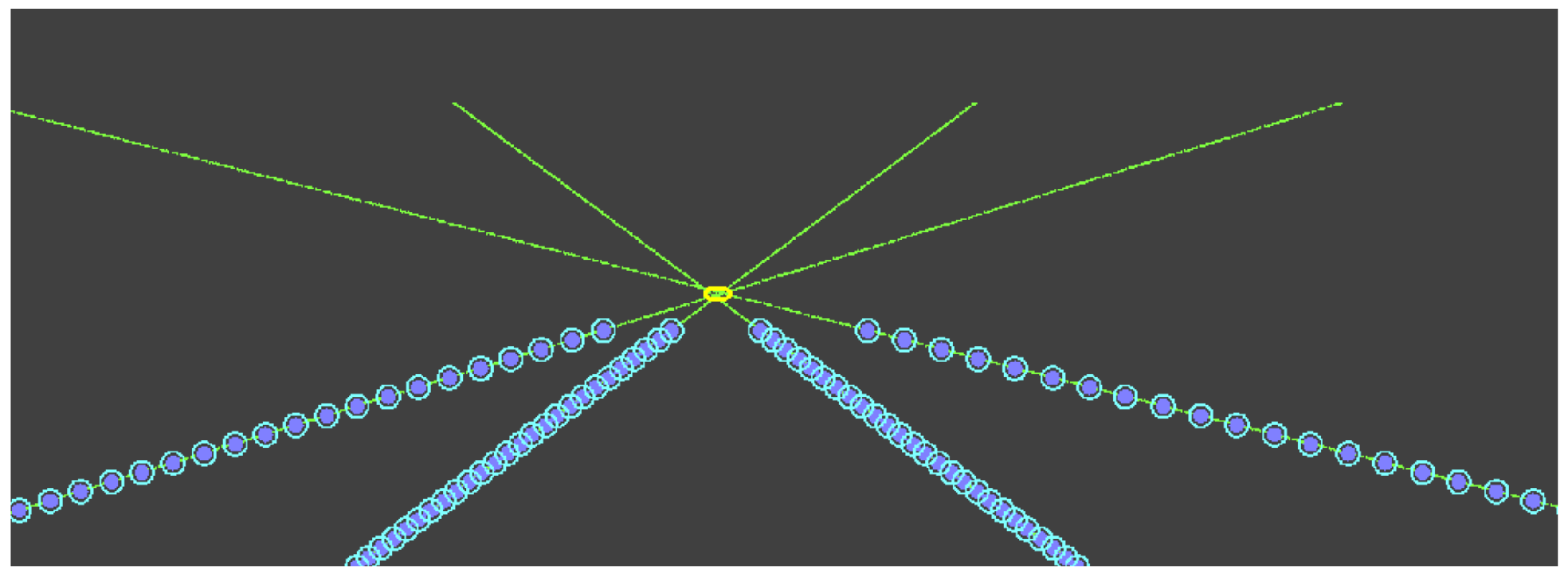}
   \includegraphics[width=0.99\linewidth]{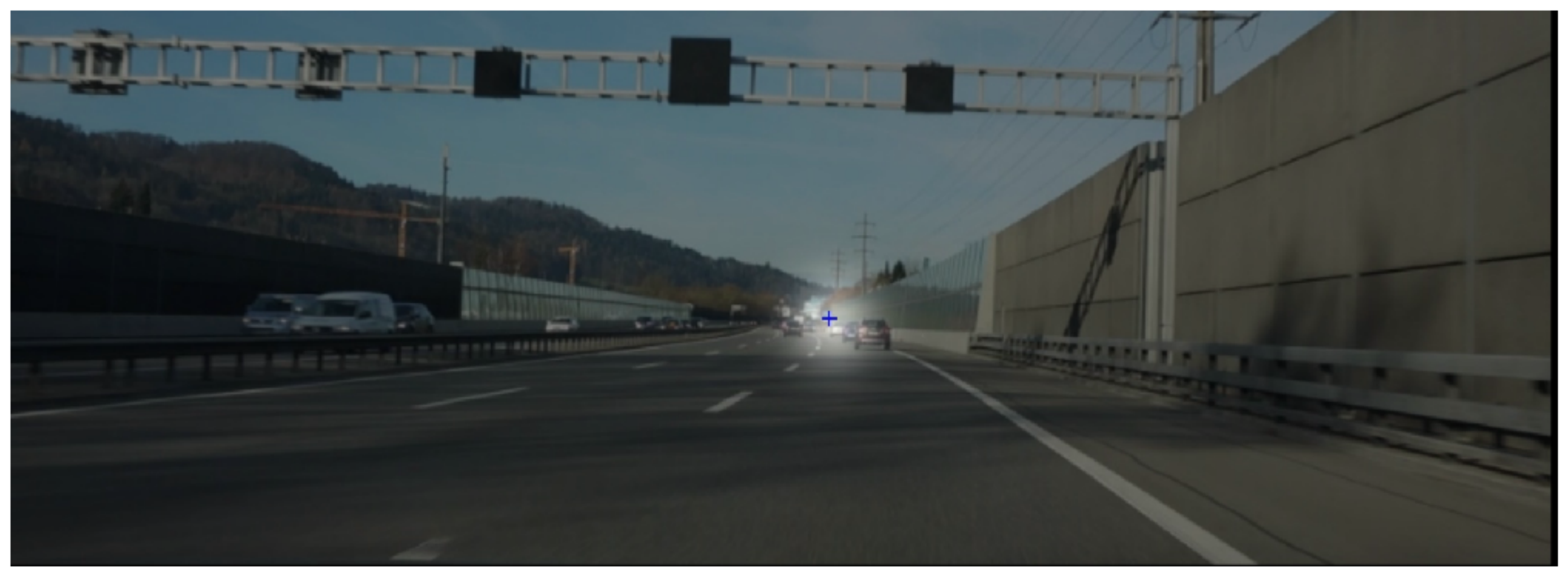}
\end{center}
   \caption{We automatically label the vanishing point (VP) exploiting lane annotations to train the VP Detector (top). At test time (bottom), the model infers probability distribution of the VP in an end-to-end manner. The output distribution is visualized as a white heat-map and the detected VP as a blue plus mark.}
\end{figure}

\section{Introduction}

There has been massive progress on the 3D scene understanding in computer vision, especially in the automotive application fields. The popular datasets such as KITTI \cite{10.1177/0278364913491297}, CityScape \cite{Cordts2016Cityscapes} and ApolloScape \cite{wang2019apolloscape} are built upon the careful camera attachment to the vehicle and calibration to obtain intrinsic and extrinsic parameters. However in the real-world scenario, obtaining external camera parameters is costly and hard to be automated. Moreover, it is sometimes required to re-align the camera when its physical attachment to the vehicle is unstable. %For instance, there could be cases in which the relationship between the camera's optical axis and the ground plane physically changes from the original condition by accidentally being moved by a passenger or a driver. 
Once the camera parameters are perturbed, the algorithms which run based on them malfunction and can not be restored. Therefore, in order to realize robust scene understanding it is essential to monitor the camera parameters after the camera is deployed. 

The vanishing points (VPs) - or point at infinity - have rich information to restore the camera parameters \cite{Hartley2000} \cite{cipollavp}. With a single VP known, camera yaw and pitch angles can be obtained. For example, when a vehicle is traveling a straight road, the VP obtained from the lane markings can determine the angle between the camera's optical axis and the vehicle's traveling direction.  When one more VP or the horizon line is detected, all the angles between the camera and the ground plane can be obtained.
The existing VP detection methods are grouped into two types: line-detection based (two-stage) and direct (end-to-end) methods. Two-stage method estimates VP from an image by detecting multiple parallel line segments \cite{10.1007/978-3-319-66709-6_2} \cite{lee2020online}. In this case the VP detection pipeline is two-stage (line detection and VP estimation) and prone to the detection error at the first stage. Moreover, when there are no hints (sufficient number of lane markers) in the image, the first stage fails and the pipeline does not work. On the other hand, the existing end-to-end detectors rely on manual annotations \cite {Lee_2017_ICCV}, which incurs annotation errors and ambiguity to achieve pixel-level accuracy for VPs.  
To see the sensitiveness of VP detection for scene recognition, let's pick an example of monocular distance estimation. When the VP is known, the distance between the camera and a point on the ground plane can be simply obtained from the y-coordinates of the point and the VP. For the CULane dataset \cite {pan2018SCNN} we use in this paper, VP error of 10 pixels along y-axis causes large distance measurement error of approx. 10m at 25m distance from the camera.

In this paper we propose a simple but effective end-to-end VP detector trained under the supervision of VP annotations automatically extracted from the lane marker annotations. By calculating geometric intersections from the lane marker annotations, we provide the model with more stable supervision without costly and unstable manual VP annotations. The VP Detector realizes accuracy of 0.0063\% average normalized Euclidean distance error compared with the VP labels, which enables practical and accurate online camera calibration. Moreover, we compare the models trained with our automatically generated labels and manual annotations, to show that our labels are more stable and geometrically consistent with the images. 

Our contributions in this paper are three-fold:
\begin{itemize}
	\item We calculate VPs exploiting lane marking annotations to obtain more accurate and stable VP position labels compared with manual annotation. Lane marker annotations are extrapolated exploiting curve fitting and the VP location is obtained as median or mean location of the multiple intersection points.
	\item An end-to-end monocular vanishing point detector coined VP Detector is developed using calculated VP labels. The VP Detector directly estimates the VP probability map and surpasses the lane-detection-first method in accuracy, especially when a lane detector fails in detecting sufficient lane markers.
	\item  By comparing manual and automatic labels, we show that manual labeling of VP positions incurs annotation errors and ambiguity, and that automatic labels can provide the detector model with more accurate and consistent supervision. \end{itemize}

\section{Related Work}

\noindent{\textbf{Vanishing Point Estimation via Line Detection.} }
The VP can be determined as the intersection of parallel lines. In the traffic environment, parallel lane markers can be utilized. Lee et al. \cite{lee2020online} firstly detect the lane markers with a lane detection model and find the VP from the lane detection results afterwards, to estimate extrinsic camera parameters. In this case the detection accuracy relies on lane detection accuracy. On the other hand, we firstly determine ground-truth VPs from lane annotations and learn them afterwards to make the pipeline lane detection free.
\\
\textbf{Lane Detection Guided by Vanishing Point.} Having the VP position as a prior is known to aid lane detection. VPGNet \cite{Lee_2017_ICCV} detects the VP and lanes simultaneously to encourage accurate lane detection. Su et al. \cite{8118305} and Ma et al. \cite{8577122} propose VP detection using v-disparity with a stereo camera to aid lane detection. Ma et al. \cite {Ma2018MultipleLD} introduce multiple lane detection utilizing the VP to estimate the road model for disparity map estimation.   \\
\textbf{Direct Vanishing Point Detection.}
Chang et al. \cite{8460499} detects VPs with CNN utilizing a VP location classification layer. Abbas et al. \cite{9022574} introduce the CNN-based VP and horizon line estimation method to obtain a homography matrix, where the network is trained on the synthetic dataset with ground truth of VPs. Lee et. al \cite{Lee_2017_ICCV}, Liu et. al \cite{liu2020heatmapbased} and Liu et al. \cite{liu2020unstructured} propose direct VP detection network trained with 2-D target map which represents manually annotated VPs. Our proposed VP detector adopts heatmap estimation method supervised by automatically labeled VPs from lane annotations.\\
\textbf{Heatmap-based Keypoint Estimation.} 
As with \cite{liu2020heatmapbased} and \cite{liu2020unstructured}, we regard VP detection as the keypoint estimation task. In human body keypoint estimation, both keypoint detection and keypoint grouping for each person have to be carried out. On the other hand, VP detection requires only detection of a single keypoint. We employ the heatmap-based keypoint detection method which is utilized in \cite{cao2017realtime} where target ground-truth keypoints are represented as a 2-D Gaussian probability distribution. The probability map is directly output from the fully-convolutional network, thus can deal with variable input resolutions.\\
%keypoints of multiple persons are detected first and grouped afterwards. \\

\begin{figure*}[t]
\begin{center}
    \centering
    \includegraphics[width=.9\linewidth]{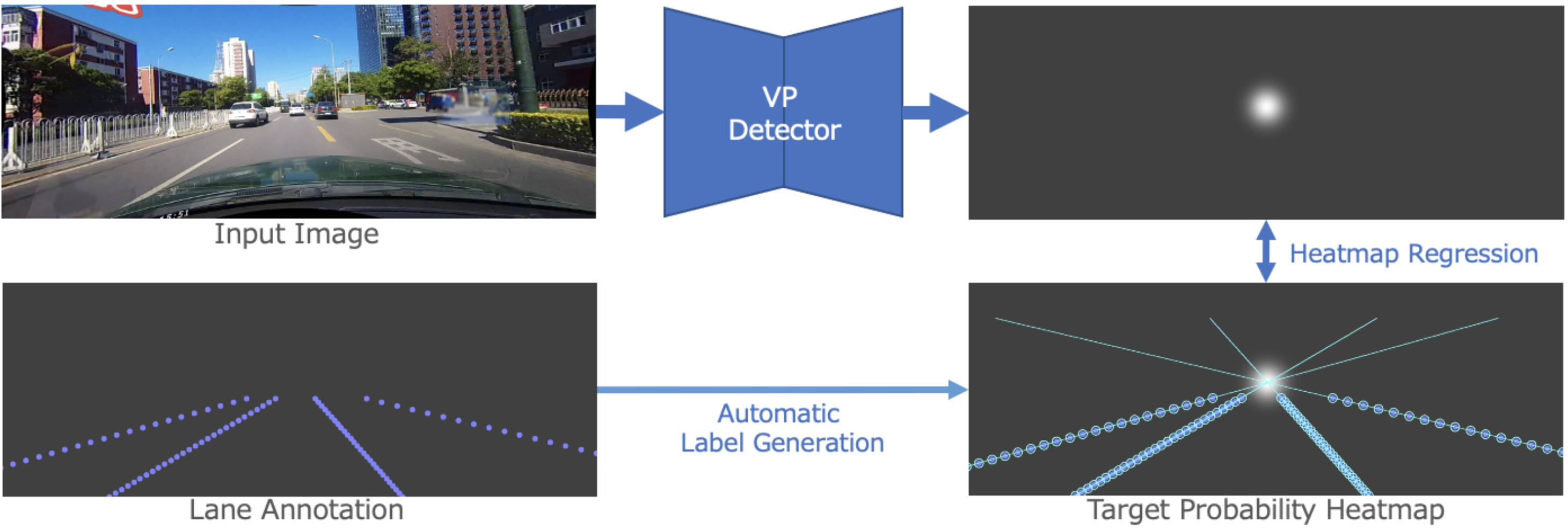}
\end{center}

   \caption{Schematic of our method. We extrapolate the lane marker annotation points via curve fitting to find the VP. The target probability heatmap has a single 2-D Gaussian distribution whose center is at the automatically labeled VP location. The convolutional neural network coined VP Detector learns the relationship between the input image and the target probability heatmap. At the inference time, VP Detector simply outputs the probability heatmap whose maximum-value position is the inferred VP.}
\label{schematic}
\end{figure*}

\section{Method}
In this section we show two components of our method: automatic label generation from lane annotations and the end-to-end VP Detector.
\subsection{Automatic Label Generation}

We extrapolate the lane annotation points to find the intersection points as the candidates of VPs. For extrapolation, we simply fit the lane annotation data in the perspective view by a \(n\)-degree polynomial to extract the VP. 
\begin{equation}
\label{polynomial}
x =  a_{n}y^{n}+a_{n-1}y^{n-1}+{...}+a_{1}y+a_{0}
\end{equation}

We employ an off-the-shelf curve fitting algorithm \cite{2020SciPy-NMeth} using least square optimization for fitting. One set of parameters corresponds to one lane marking which consists of multiple annotation points. 
The definition of VP varies in the driving scenes especially where the road is not straight. To this end, we aim to label two types of VPs with curve fitting: \(VP_{road}\) where the lane markings supposedly go and \(VP_{straight}\) where the parallel lane markings at the close range meet.  \(VP_{road}\) is obtained by conducting curve fitting including the far range. %Higher-degree polynomials can give more plausible but more unstable approximation of the far-range lanes. 
On the other hand, \(VP_{straight}\) is obtained by picking the lane marking annotations at the close range and employing 1-degree fitting (1D-close fitting). \(VP_{straight}\) is not affected by the curves and the VP represents the direction of the approximately straight lane markers close to the ego-vehicle, which we consider more practical for camera calibration.

\begin{figure}[htb]
\centering
  \begin{subfigure}[b]{.508\linewidth}
    \centering
    \includegraphics[width=.99\linewidth]{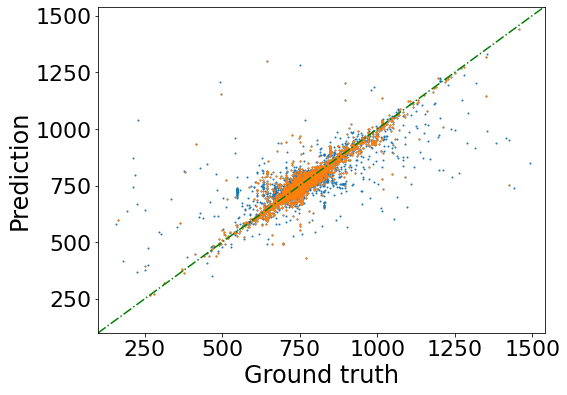}
    \caption{}
  \end{subfigure}
  \begin{subfigure}[b]{.481\linewidth}
    \centering
    \includegraphics[width=.99\linewidth]{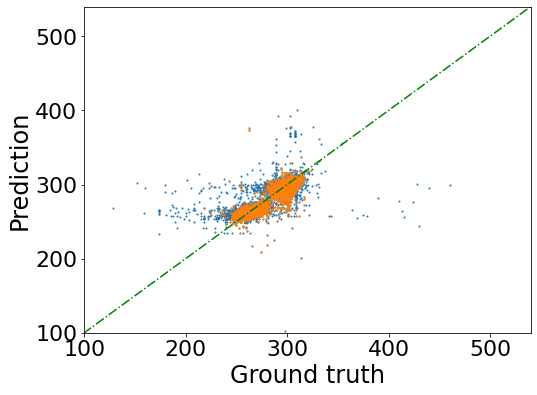}
    \caption{}
  \end{subfigure}\\% 
  \begin{subfigure}[b]{.95\linewidth}
    \centering
    \includegraphics[width=.99\linewidth]{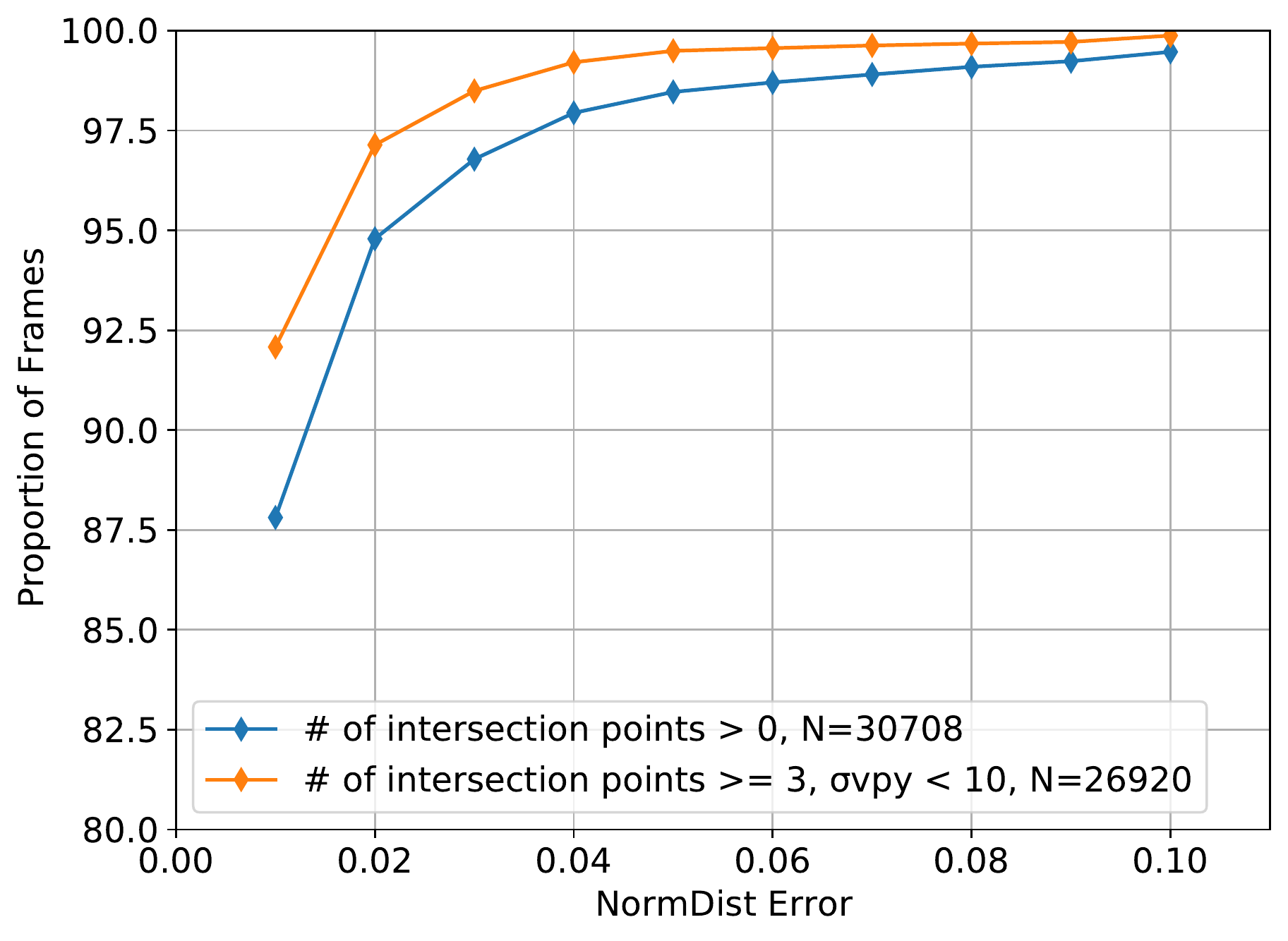}
    \caption{}
  \end{subfigure}\\%  
  \caption{Evaluation results of our VP Detector on the CuLane test dataset (30,708 successfully labeled frames out of 34,679). (a) (b) Comparison between predicted VP and ground truth along (a): x-axis and (b): y-axis. (c) Cumulative error distribution of normalized Euclidean distance. Orange and blue dots represent different \textbf{test data} subsets - orange dots : \(N_{int}\geq3\), \(\sigma_{vpy} < 10\) (\(N=26920\)) and blue dots : \(N_{int}>0\) (\(N=30708\)).}
  \label{fig:first_result}
\end{figure}

\begin{figure}[t]
\begin{center}
   \includegraphics[width=0.99\linewidth]{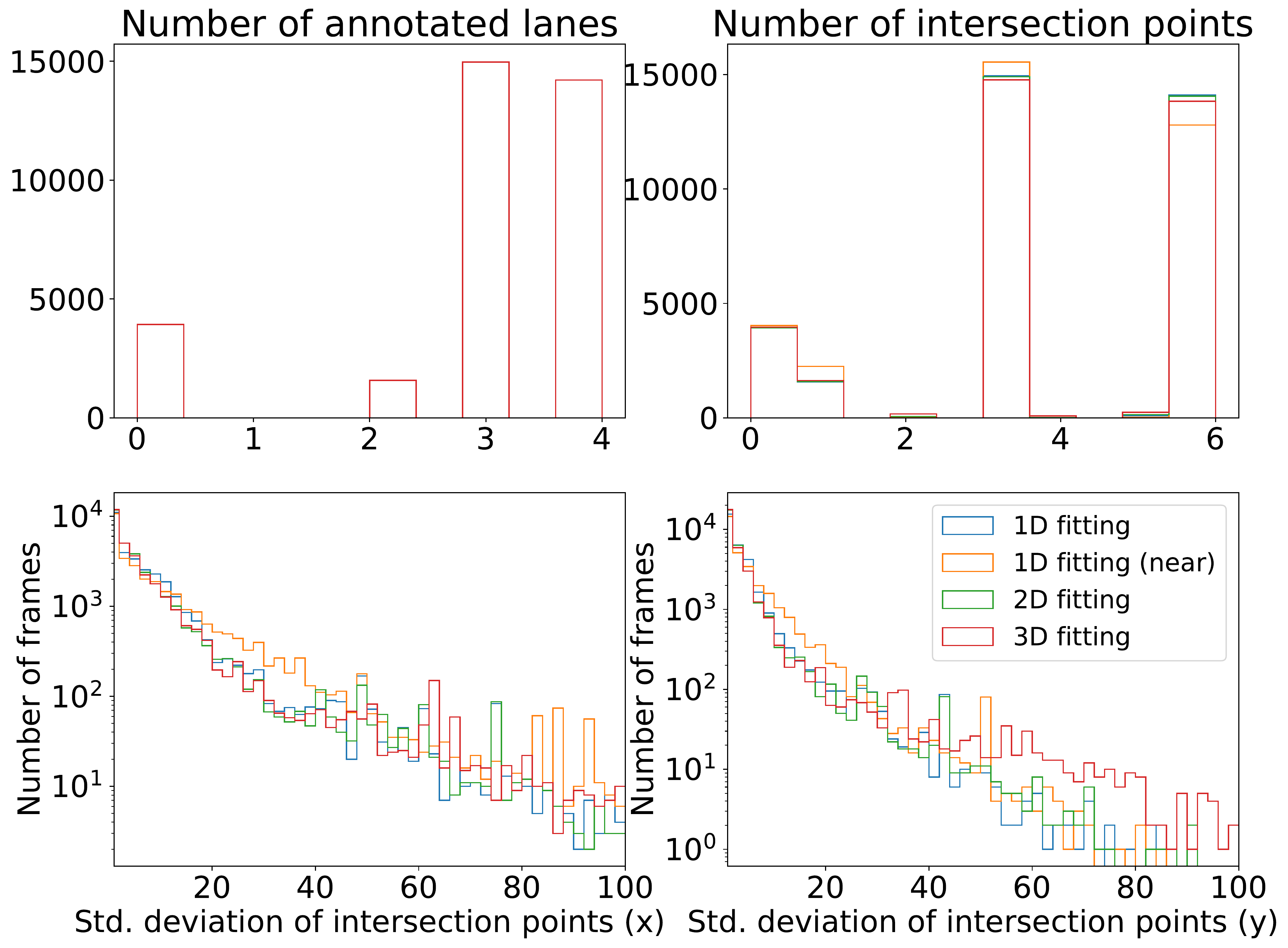}
\end{center}
  \caption{Label generation results on the CULane test set that consists of 34,680 frames. top left: number of annotated lanes, which varies from 0 to 4. top right: histogram of number of intersection points. bottom: standard deviation of lane intersection points along x-axis (left) and y-axis (right). }
\label{fig:label_stats}
\end{figure}

\begin{figure*}[htb]
\centering
  \begin{subfigure}[b]{.32\linewidth}
    \centering
    \includegraphics[width=.99\linewidth]{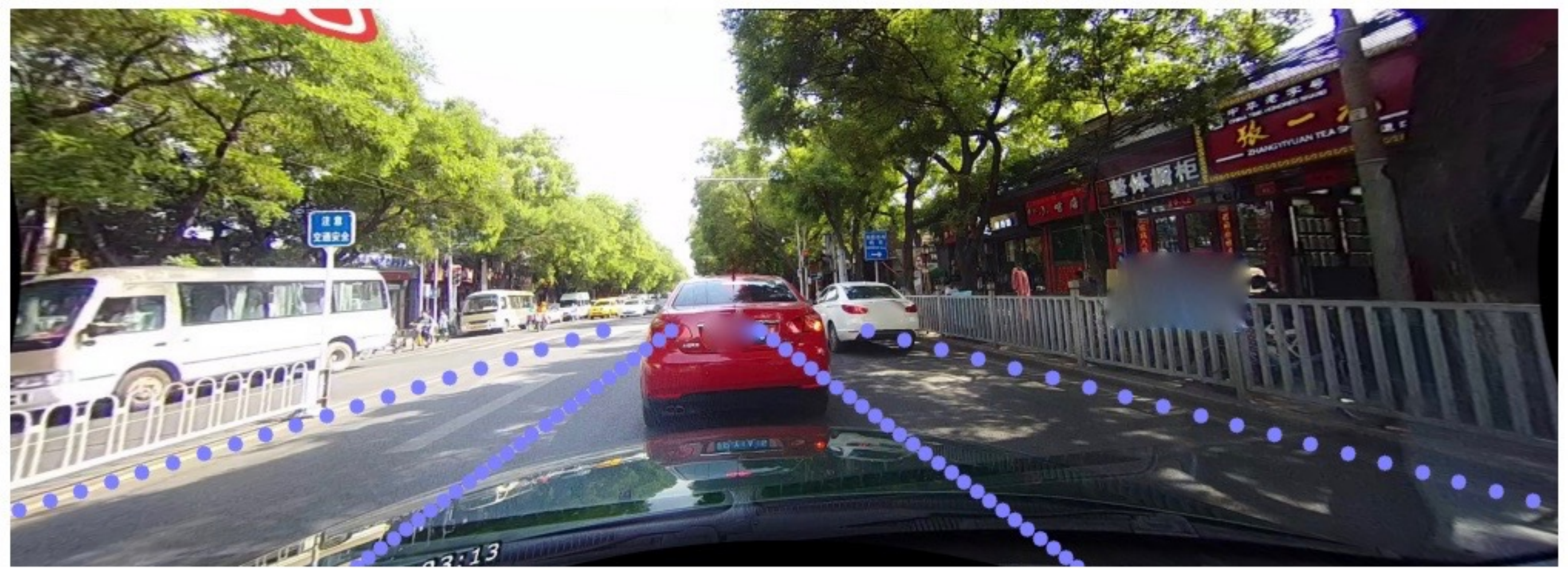}
  \end{subfigure}
  \begin{subfigure}[b]{.32\linewidth}
    \centering
    \includegraphics[width=.99\linewidth]{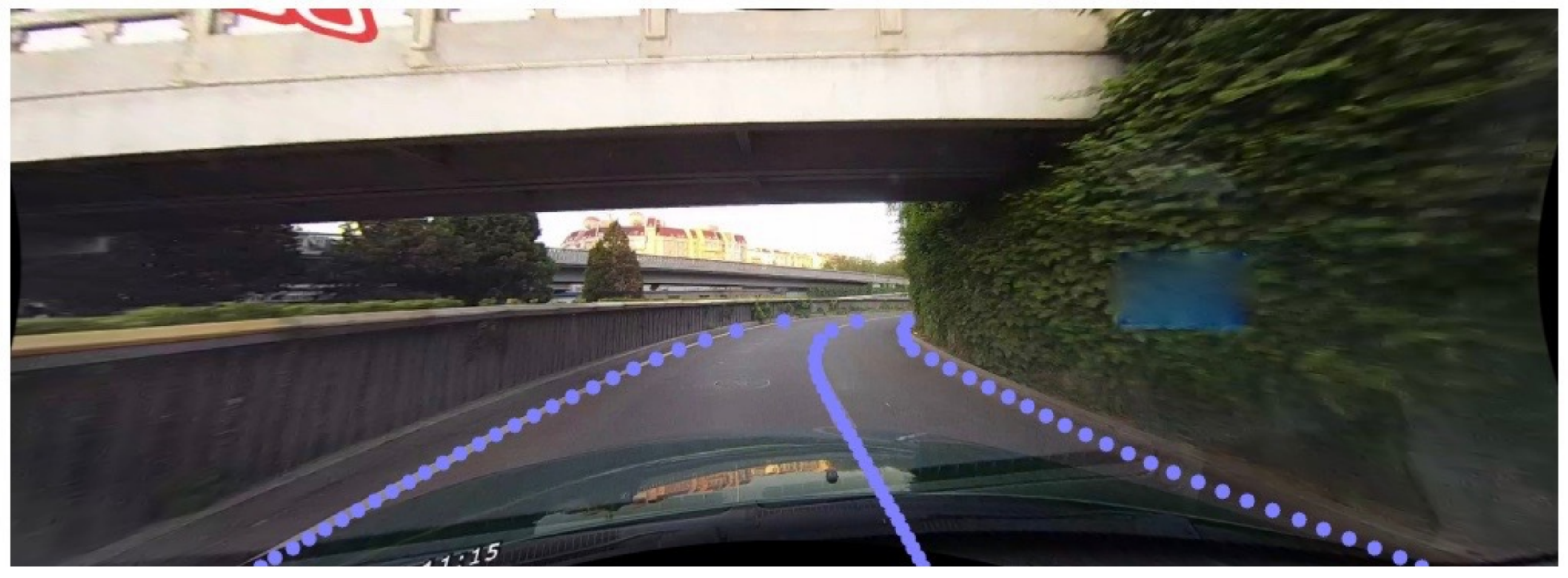}
  \end{subfigure}
  \begin{subfigure}[b]{.32\linewidth}
    \centering
    \includegraphics[width=.99\linewidth]{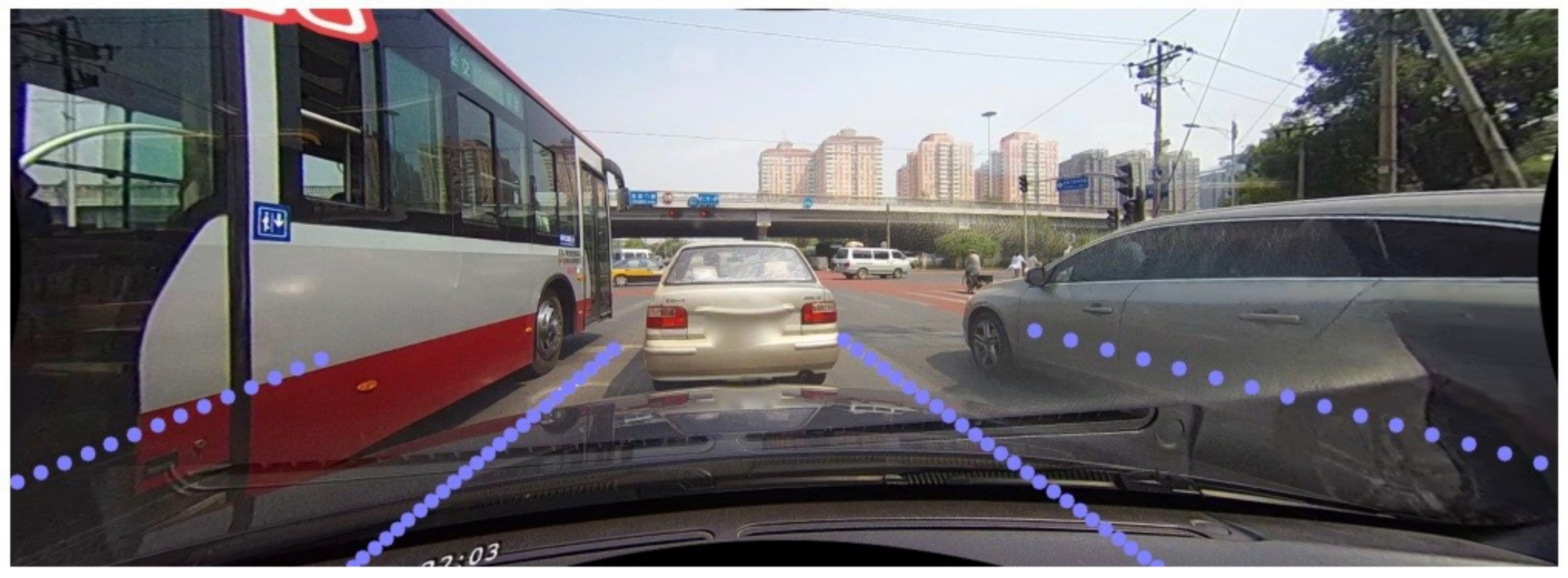}
  \end{subfigure}\\%  
  
  \begin{subfigure}[b]{.32\linewidth}
    \centering
    \includegraphics[width=.99\linewidth]{data/1D_curvefit_driver_100_30frame_05251502_0428_01230.pdf}
  \end{subfigure}
  \begin{subfigure}[b]{.32\linewidth}
    \centering
    \includegraphics[width=.99\linewidth]{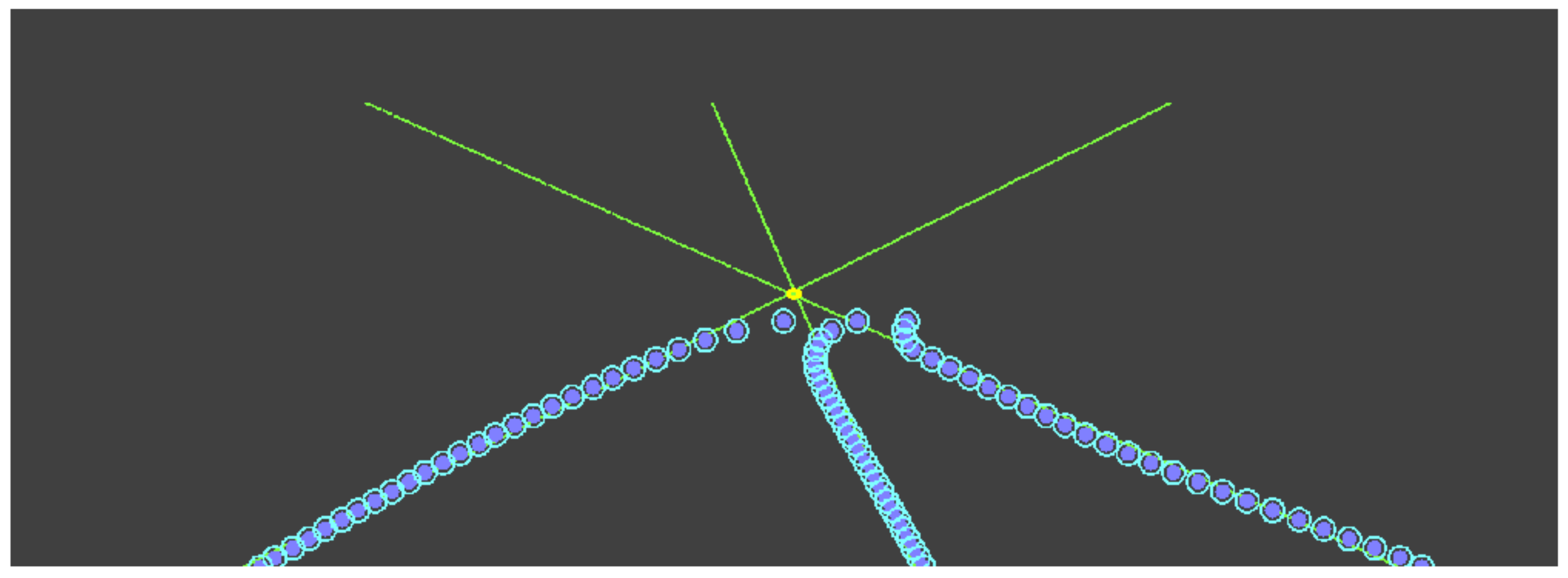}
  \end{subfigure}
  \begin{subfigure}[b]{.32\linewidth}
    \centering
    \includegraphics[width=.99\linewidth]{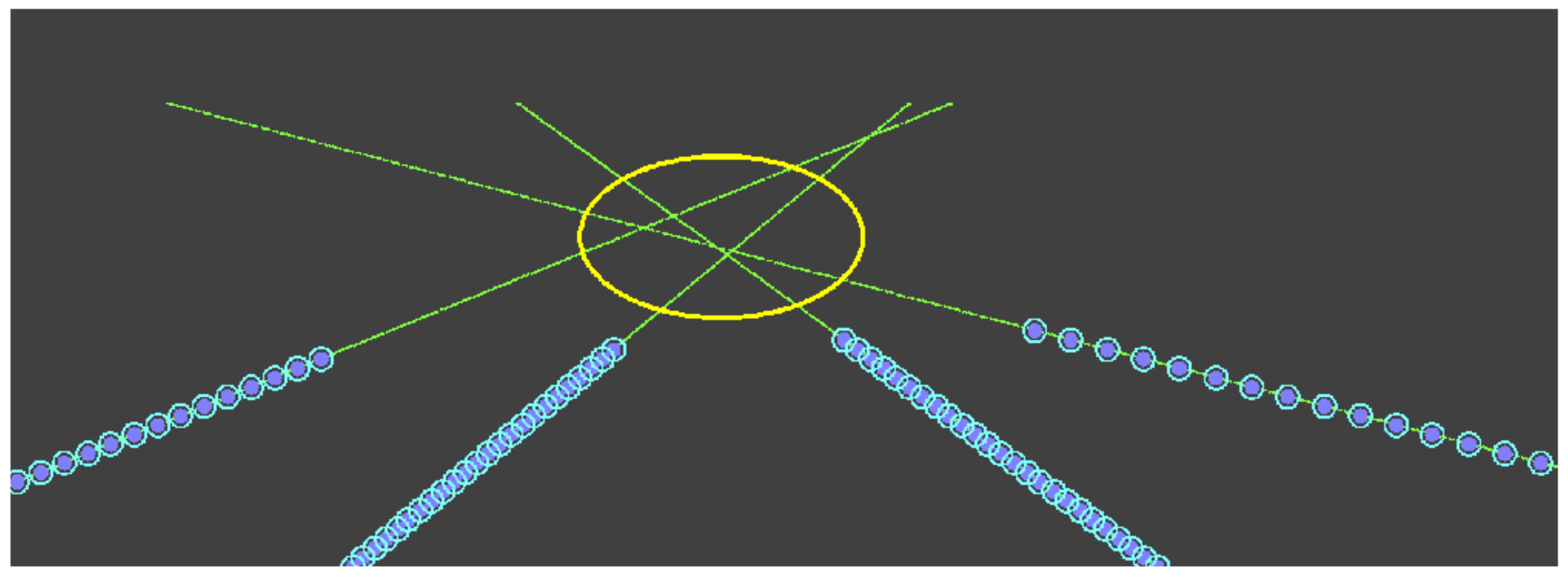}
  \end{subfigure}\\%  
  
    \begin{subfigure}[b]{.32\linewidth}
    \centering
    \includegraphics[width=.99\linewidth]{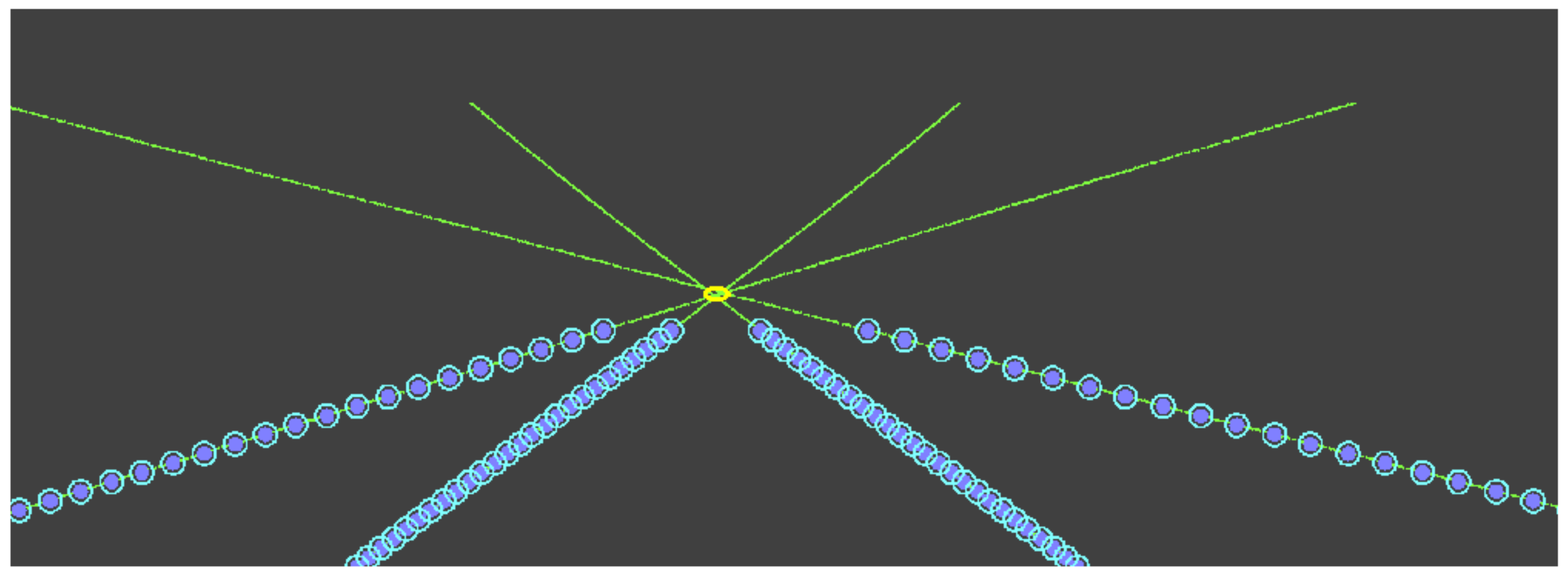}
  \end{subfigure}
  \begin{subfigure}[b]{.32\linewidth}
    \centering
    \includegraphics[width=.99\linewidth]{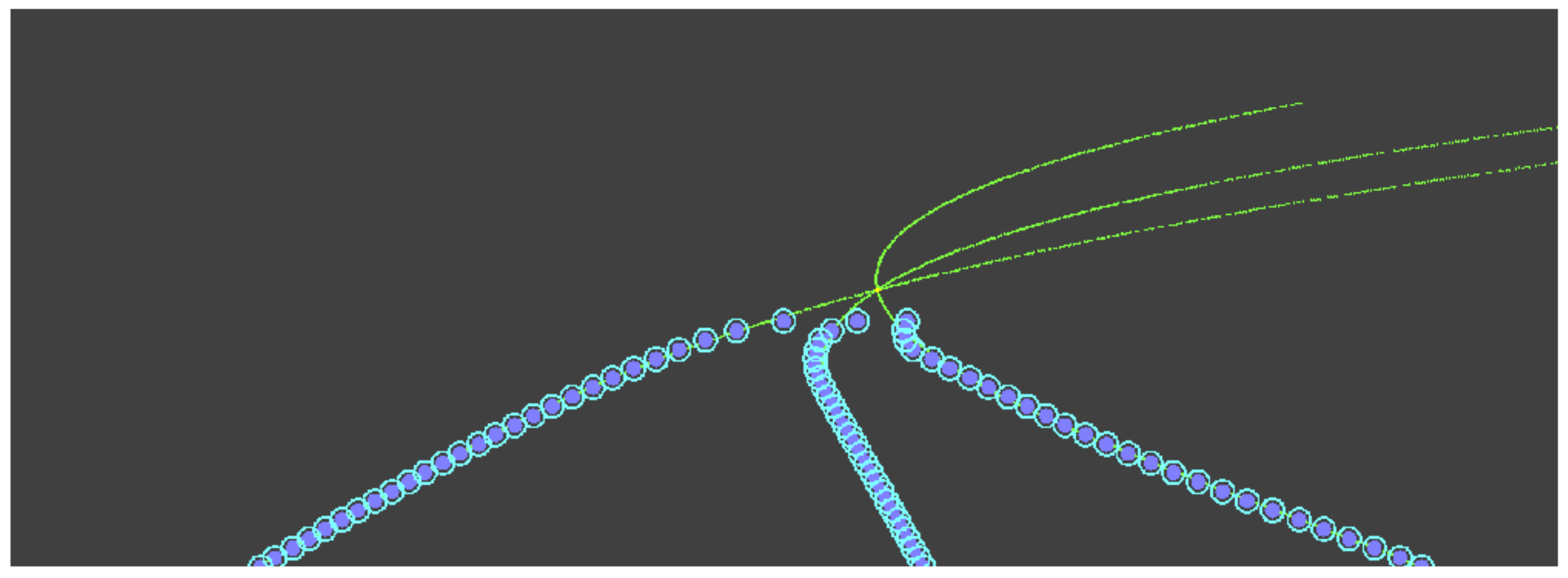}
  \end{subfigure}
  \begin{subfigure}[b]{.32\linewidth}
    \centering
    \includegraphics[width=.99\linewidth]{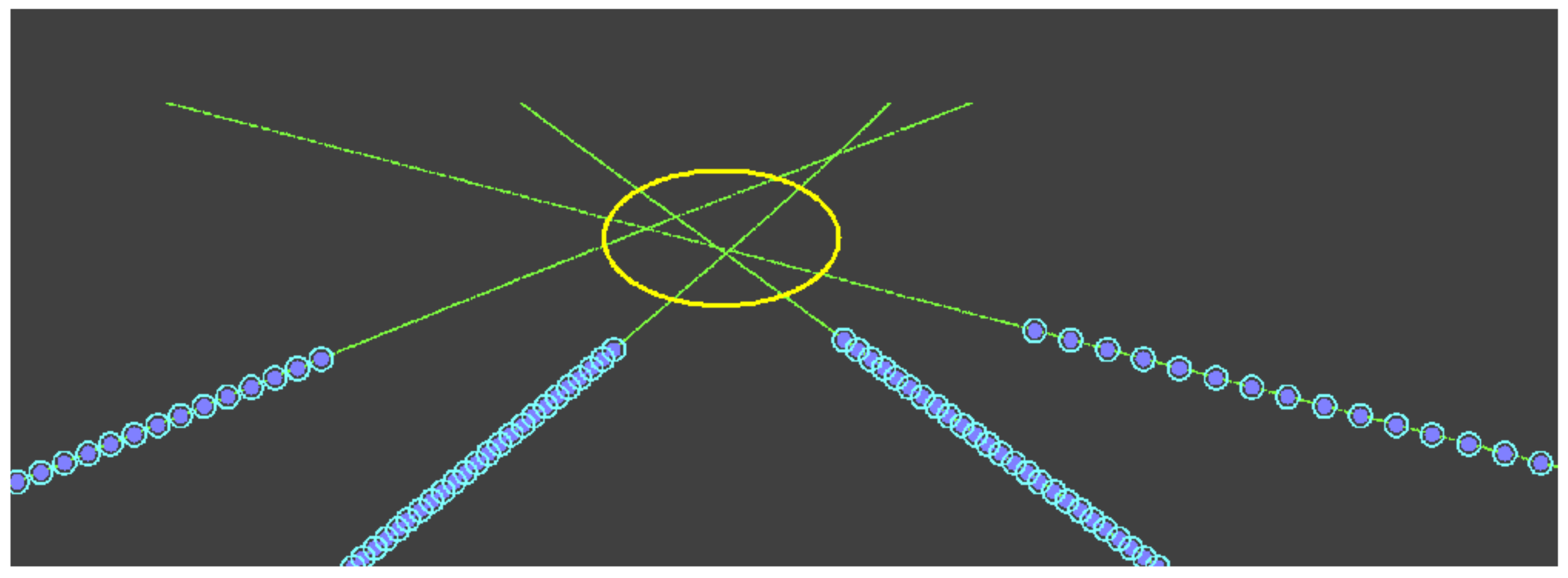}
  \end{subfigure}\\%  
  
    \begin{subfigure}[b]{.32\linewidth}
    \centering
    \includegraphics[width=.99\linewidth]{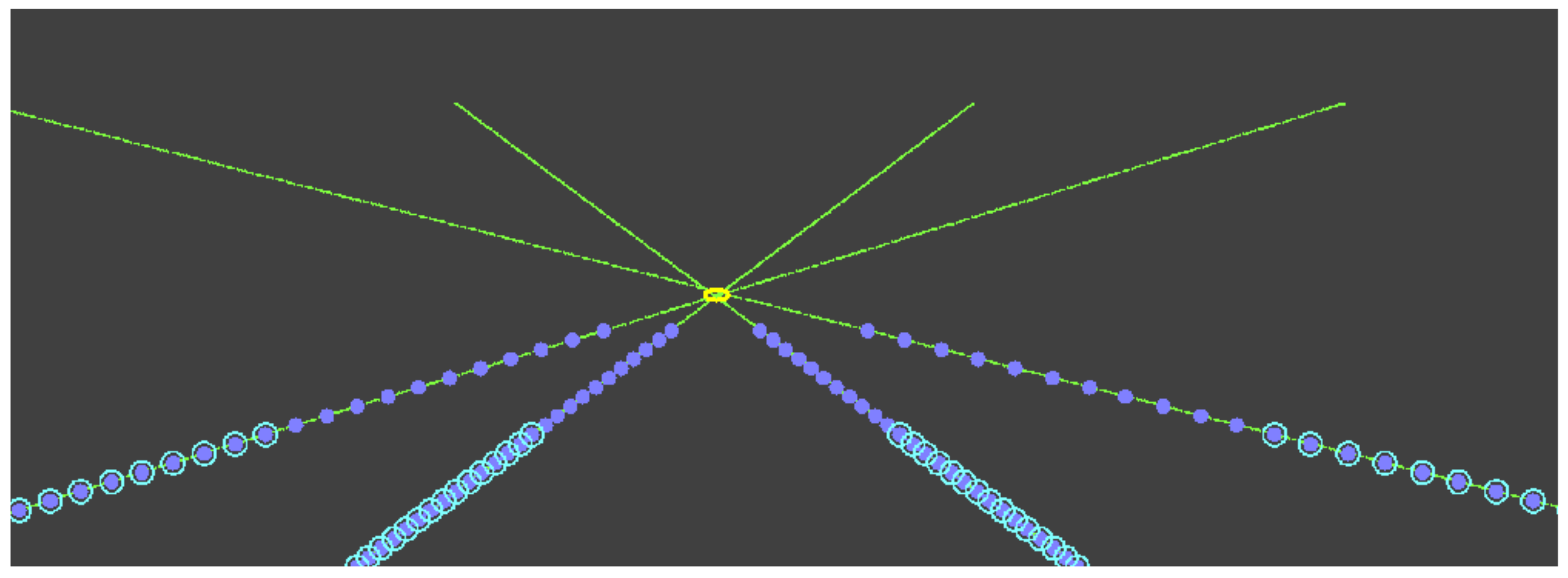}
    \caption{}
  \end{subfigure}
  \begin{subfigure}[b]{.32\linewidth}
    \centering
    \includegraphics[width=.99\linewidth]{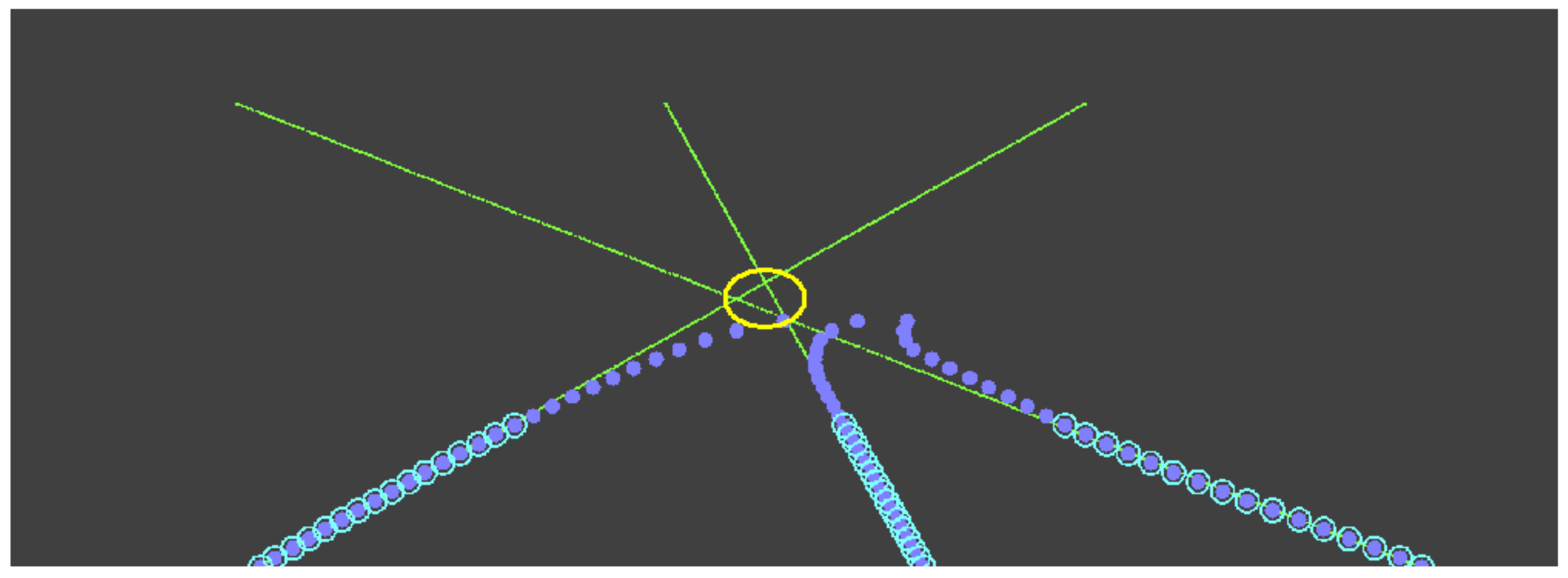}
    \caption{}
  \end{subfigure}
  \begin{subfigure}[b]{.32\linewidth}
    \centering
    \includegraphics[width=.99\linewidth]{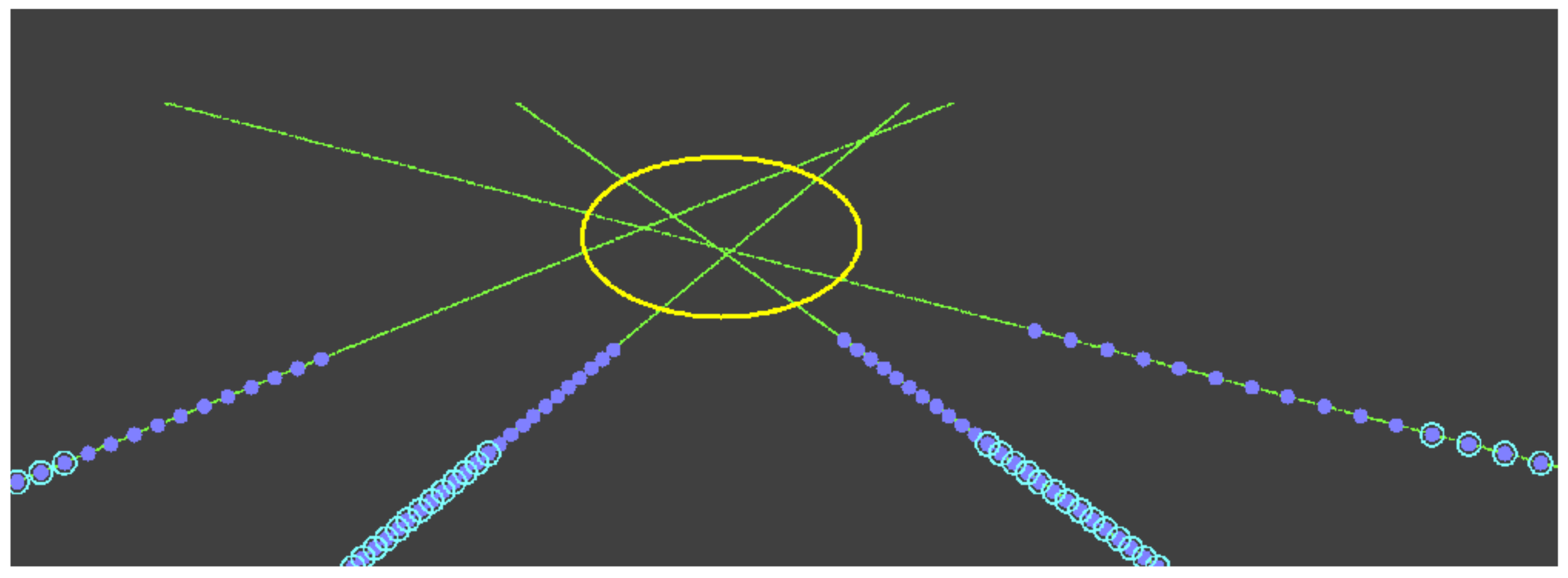}
    \caption{}
  \end{subfigure}\\%  
   \caption{Label generation on the CuLane dataset. The lane annotation points and extrapolation curves are visualized. The center and the size of the yellow oval at the intersection area represent the median and standard deviation of the intersection points. The lane annotation points employed for fitting are highlighted by white circles. From the first row, images with the lane marking annotations from the CuLane dataset, results of 1D, 3D and 1D-close fitting settings are shown. The column (a) shows an example with four lane markers, column (b) curved three-lane frame and (c) a failure case where the left-most lane annotation is inconsistent with the others. Best viewed in color.}\label{fig:find_vp}
\end{figure*}
\subsection{End-to-end Vanishing Point Detector}\label{section:vpdetector}

The schematic of our VP Detector is shown in Fig. \ref{schematic}. We employ a CNN with an hourglass \cite{870a37a4f4e24a66bce3afcddaadf871} (or U-Net \cite{RFB15a}) architecture as a VP detector model. The VP detection is equivalent to the keypoint detection using heatmaps, therefore there are plenty of network choices. The output of the network is the single-channel heatmap with the same resolution as the input. The training target is a 2-D heatmap that has a single VP as 2-D Gaussian distribution whose max value is 1:
\begin{equation}
H(x, y) = A\exp\left(-\frac{x^2+y^2}{2\sigma^{2}}\right)\,
\end{equation}
where \(\sigma\) controls the width of the distribution. 
In the experiment section we compare 1) static setting where \(\sigma\) has a fixed value and \(A=1\), 2) dynamic setting where \(\sigma\) represents the standard deviation of the multiple VP candidates and \(A=1\), and 3) dynamic setting with \(A=1/\sigma\sqrt {2\pi}\), the regular Gaussian distribution.
At prediction time, the maximum value and its location are extracted from the output probability map as a confidence value and the VP. 

\section{Experiments}
In this section, we demonstrate our automatic label generation, training and evaluation of our VP Detector, showing that our method is advantageous over manual annotations and the lane-detection based method in VP detection accuracy.
\subsection{Datasets}
\noindent
\textbf{Lane Annotations.} We adopt the CULane dataset\footnote{\url{https://xingangpan.github.io/projects/CULane.html}} \cite {pan2018SCNN} contains 88,880 train frames, 9,675 validation frames, and 34,680 test frames with lane point annotations. In this paper VP Detector is evaluated on the test frames and the validation split is not utilized. 

VPGNet-DB-5ch\footnote {\url{https://github.com/SeokjuLee/VPGNet}} \cite{Lee_2017_ICCV} is the dataset with 17-class road marking labels and VP annotations. Since lane marking annotations are given as segmentation masks in the VPGNet-DB dataset, we have extracted lane centerlines from the masks and used them for curve fitting: The center point of each lane marking is calculated from the left and right edges of the lane segmentation mask for every horizontal line at 5 pixel intervals. The extracted points are grouped as a line for each lane marking utilizing the masks. We have split the dataset into train and test by the recording date. Only the lane markers that are longer than 50 pixels along the y-axis are employed.

As shown in Table \ref{tab:nlanes}, the frames vary in the number of annotated lanes. Since our label generation requires more than two lanes to extract intersection points as VP candidates, the frames with less than one lane (\(\sim12\%\) for CULane dataset) are not employed. \\
\subsection{VP Labeling Results}
\noindent{\textbf{Curve Fitting Settings.} Label generation is conducted using the polynomial eq. \ref{polynomial} with degrees of 1, 2 and 3 (1D, 2D and 3D settings). For 1-d fitting, we add the 1D-close setting where the lane annotations at close range are employed for fitting to label \(VP_{straight}\). More specifically, we pick the points whose y coordinates are more than the line 100 pixels below the top-most annotations.} \\
\textbf{Label Generation Results.} Fig. \ref{fig:label_stats} shows the fitting results for the test frames. For the frames with 2, 3 and 4 lane annotations, ideally \(_{2}C_{2}=1\), \(_{3}C_{2}=3\) and \(_{4}C_{2}=6\) intersection points are calculated respectively. The number of failure cases where the ideal number of intersection points is obtained differ among curve fitting settings as shown in the top left of Fig. \ref{fig:label_stats}, however the frequency is low. The standard deviation of lane intersection points (bottom of Fig. \ref{fig:label_stats}) is less than 20 pixels in the majority of cases. Fig. \ref{fig:find_vp} shows the visual examples of the label generation. For the straight lane case (left column), the results are similar among the fitting conditions. For the curved lanes, fitting with higher degrees follows the annotations more than the 1D setting.  The failure case (right column) has large deviations of the intersection points. The inconsistency of the lines is caused by the incorrect annotations on the leftmost lane marking that is occluded by a bus. The automatic labeling results can be filtered by means of thresholding on deviation of the intersection points.
\begin{table}
\begin{center}
\begin{tabular}{|l|c|c|c|c|c|}
\hline
\# of lanes & CULane  & CULane& VPG  & VPG\\
 & train  & test & train & test\\
\hline
0 & 10459 & 3927& 1651& 978\\
1 & 10 & 0& 3052 & 1020\\
2 & 4795 & 1579& 5617  & 1266\\
3 & 37757 & 14968 & 4272& 974\\
4 & 35859& 14206 & 1299 & 419\\
\(\geq\)5 & 0 & 0 & 196 & 92 \\
\hline

Total & 88880 & 34680 & 16087 & 4749\\
\hline
More than& 78411  & 30753& 11384 & 2751 \\
2 lanes& (88.2\%) &  (88.7\%)& (70.8\%) & (57.9\%) \\
\hline
\end{tabular}
\end{center}
\caption{Number of annotated lanes available for automatic VP labeling in the CULane dataset and the VPG-DB-5ch dataset.}
\label{tab:nlanes}
\end{table}

\subsection{Training and Evaluation}

As a VP detector model, we employ ERFNet \cite {Romera2018ERFNetER} for comparison with \cite{liu2020heatmapbased}. 
Horizontal flip and vertical shift augmentation are applied during training with probabilities of 0.5 and 0.5 respectively. The target heatmap is a Gaussian distribution whose peak is at the median point of the lane intersection points and the deviation is a fixed value (\(\sigma = 16\)) as the baseline condition. As a loss function we adopt mean squared error (MSE) between predicted and target heatmaps.
The network is trained from scratch for 300,000 iterations with batch size as 16, which takes 124 hours for input resolution of \((h, w) = (295, 820)\) on a single T4 GPU. 
Adam \cite{DBLP:journals/corr/KingmaB14} is used for optimization with initial learning rate as 0.001, which is dropped to 0.0001 after 240,000 iterations.\\
\textbf{Shift Augmentation}\label{section:shiftaugmentation}
The lane datasets contain a limited number of camera pose settings because a camera is fixed in each driver sequence. Thus the VP detector does not generalize well. To this end, we apply simple shift augmentation during training to mimic the camera pitch perturbation.
Let \(vp_{y}\) be a relative y-coordinate of the ground-truth VP. The training image and target heat-map are shifted by a random value which ranges from \(h_{edge}-vp_{y}\) to \(1-vp_{y}-h_{edge}\). As a result, the target \(vp_{y}\) distribution ranges from \(h_{edge}\) to \(1-h_{edge}\) homogeneously.
There is more diversity regarding \(vp_{x}\) than \(vp_{y}\) in the dataset since the yaw direction of the vehicle frequently changes during a driving sequence. Thus in this paper we employ augmentation only for \(vp_{y}\).\\
\textbf{Evaluation Metrics.}
We employ mean absolute error (MAE) along the x-axis and the y-axis and normalized distance (NormDist) as the evaluation metrics. NormDist employed by \cite{liu2020heatmapbased} is the ratio of the absolute Euclidean distance between predicted VP and the ground truth and the diagonal of the input image. For the CULane and VPG-DB-5ch datasets, \(NormDist=0.01\) corresponds to a VP error of 17 and 8 pixels respectively. 

\subsection{Evaluation on CULane dataset}
\begin{figure*}[t]
\begin{center}
  \begin{subfigure}[b]{.33\linewidth}
    \centering
    \includegraphics[width=.99\linewidth]{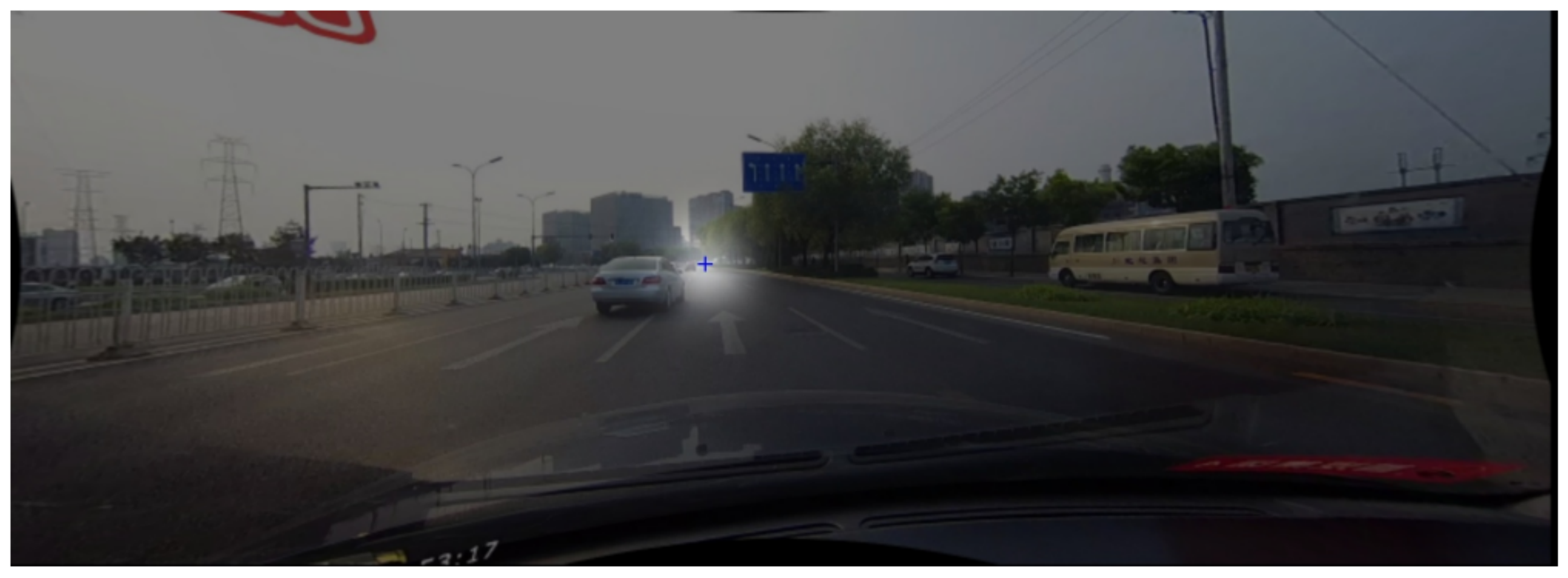}
  \end{subfigure}
  \begin{subfigure}[b]{.33\linewidth}
    \centering
    \includegraphics[width=.99\linewidth]{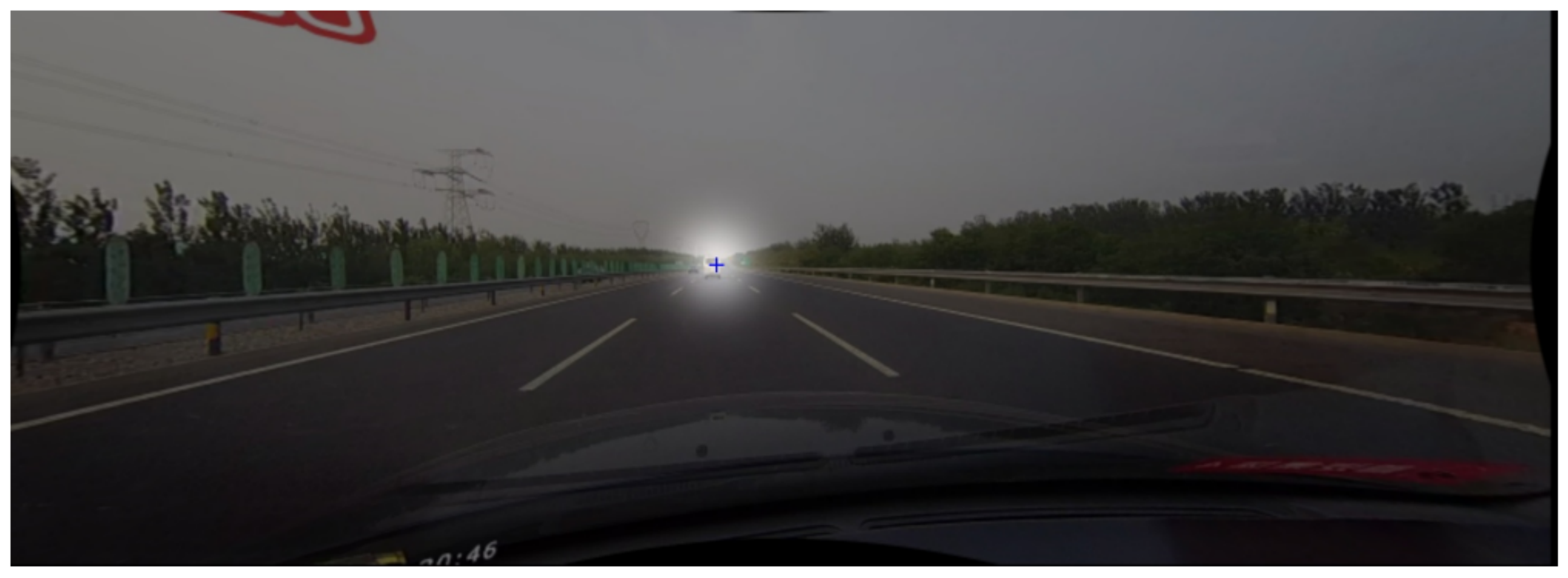}
  \end{subfigure}
  \begin{subfigure}[b]{.33\linewidth}
    \centering
    \includegraphics[width=.99\linewidth]{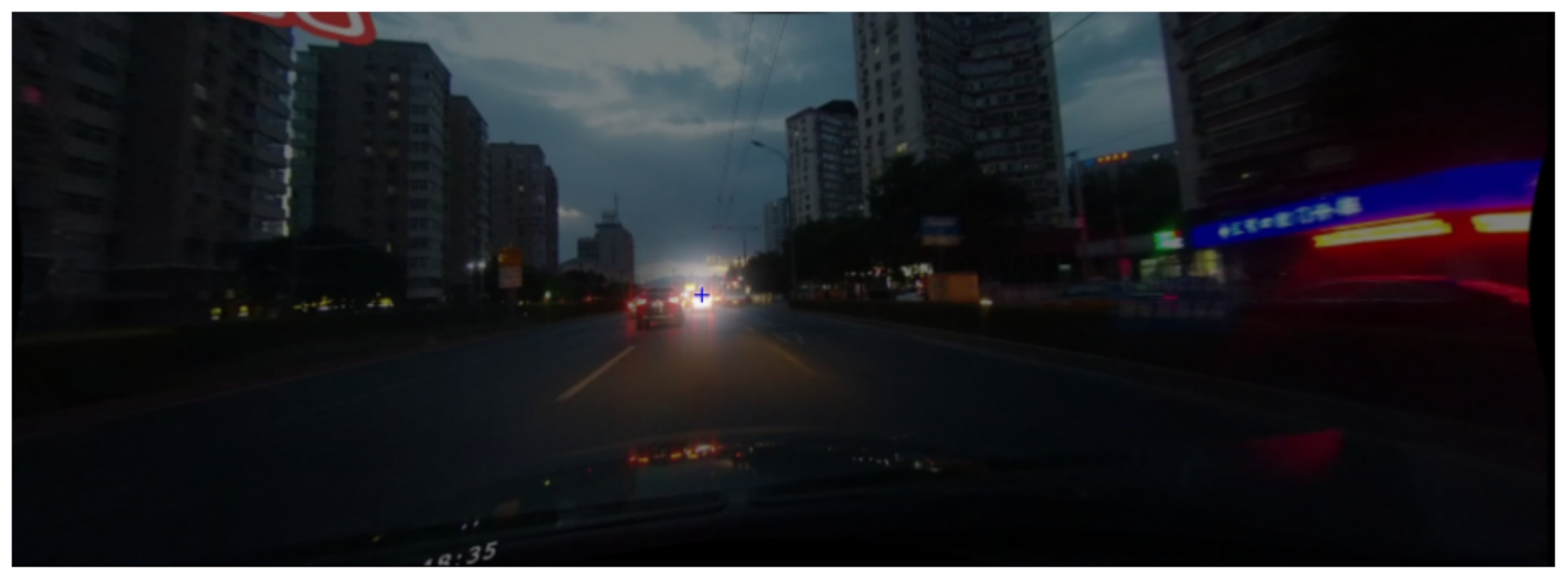}
  \end{subfigure}\\
  \begin{subfigure}[b]{.33\linewidth}
    \centering
    \includegraphics[width=.99\linewidth]{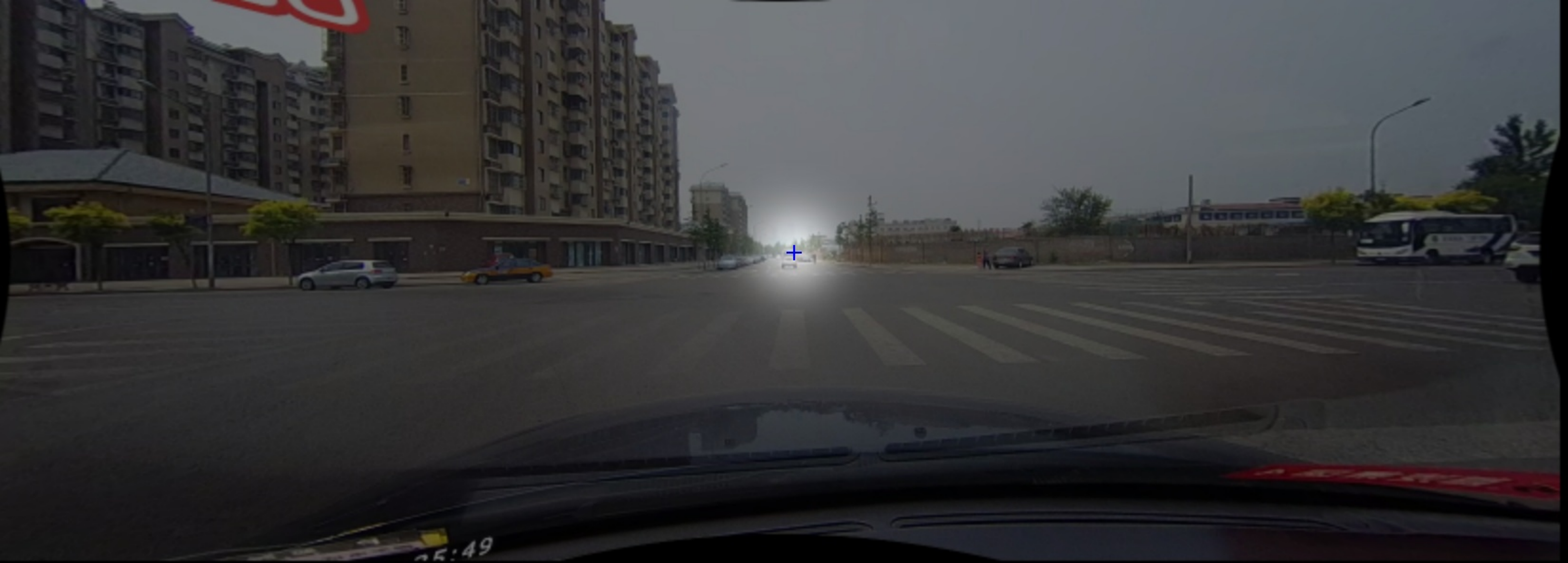}
  \end{subfigure}
  \begin{subfigure}[b]{.33\linewidth}
    \centering
    \includegraphics[width=.99\linewidth]{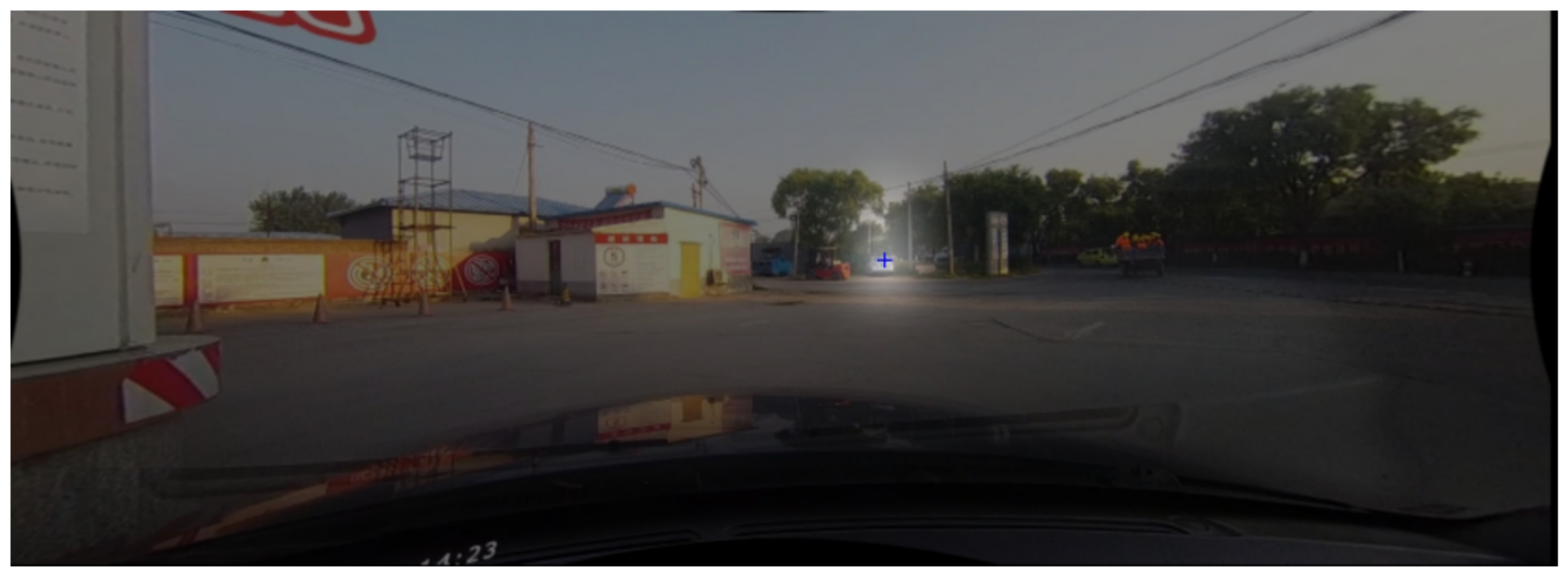}
  \end{subfigure}
  \begin{subfigure}[b]{.33\linewidth}
    \centering
    \includegraphics[width=.99\linewidth]{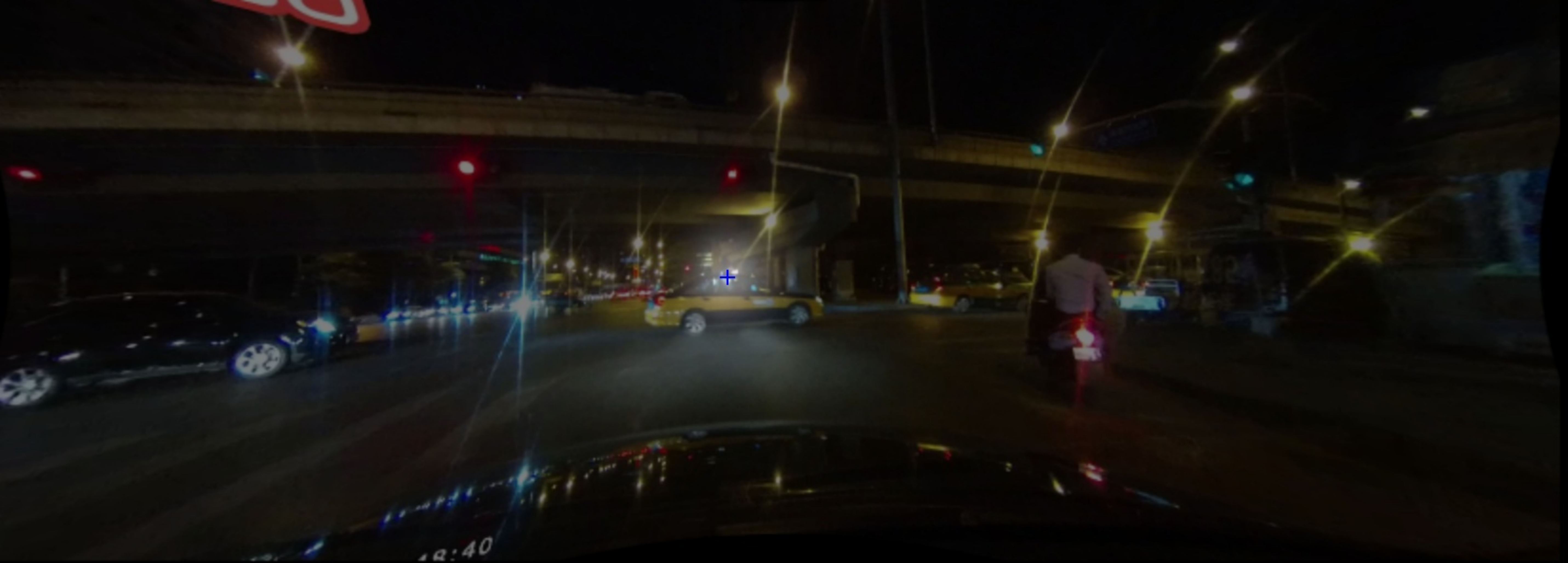}
  \end{subfigure}\\
    \begin{subfigure}[b]{.33\linewidth}
    \centering
    \includegraphics[width=.99\linewidth]{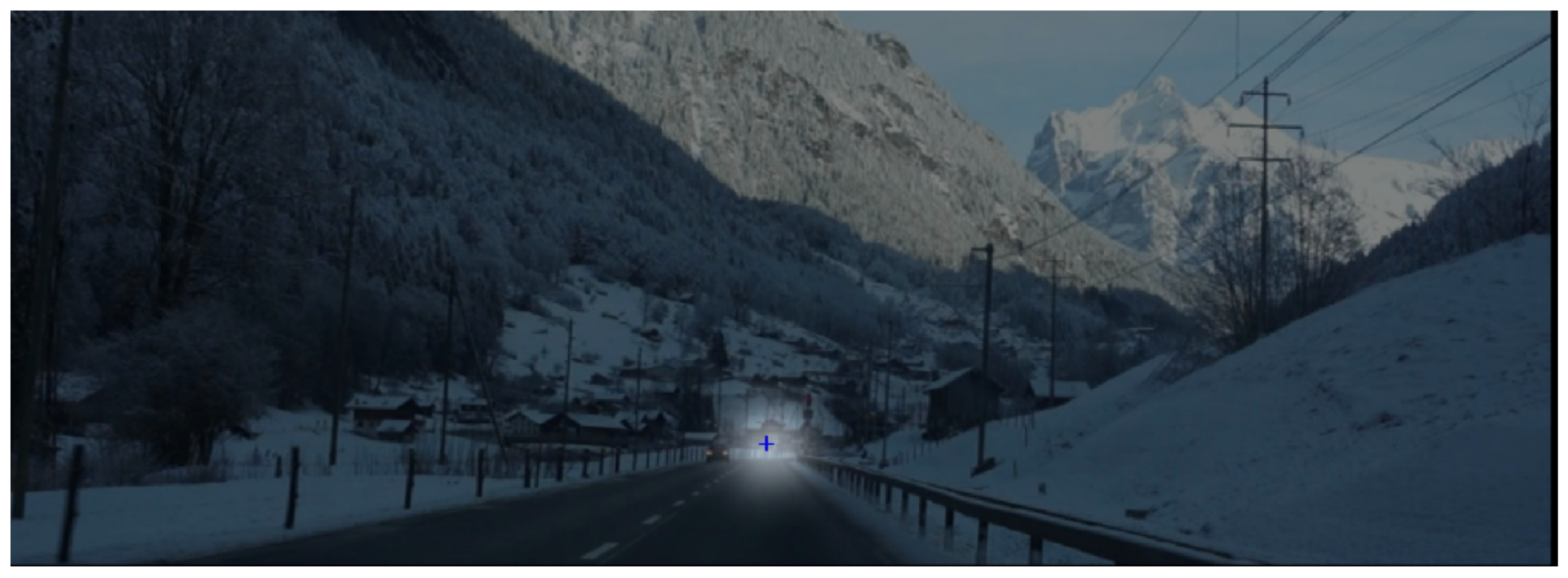}
  \end{subfigure}
  \begin{subfigure}[b]{.33\linewidth}
    \centering
    \includegraphics[width=.99\linewidth]{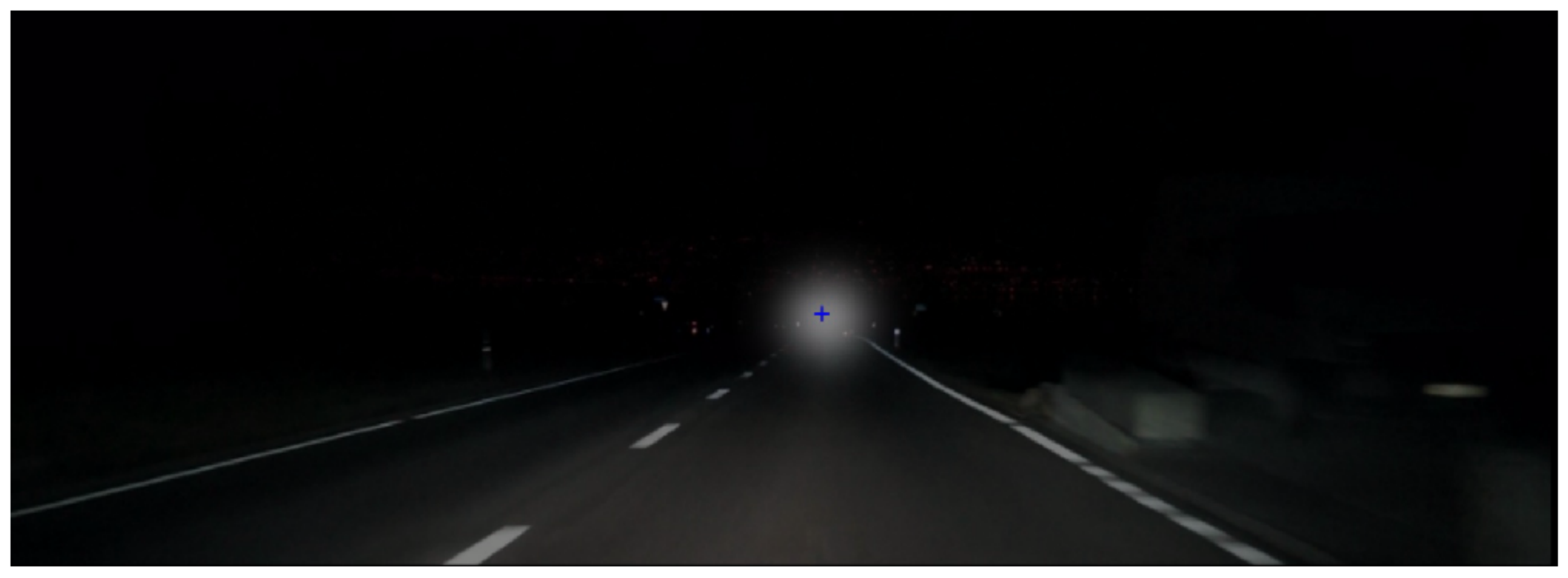}
  \end{subfigure}
  \begin{subfigure}[b]{.33\linewidth}
    \centering
    \includegraphics[width=.99\linewidth]{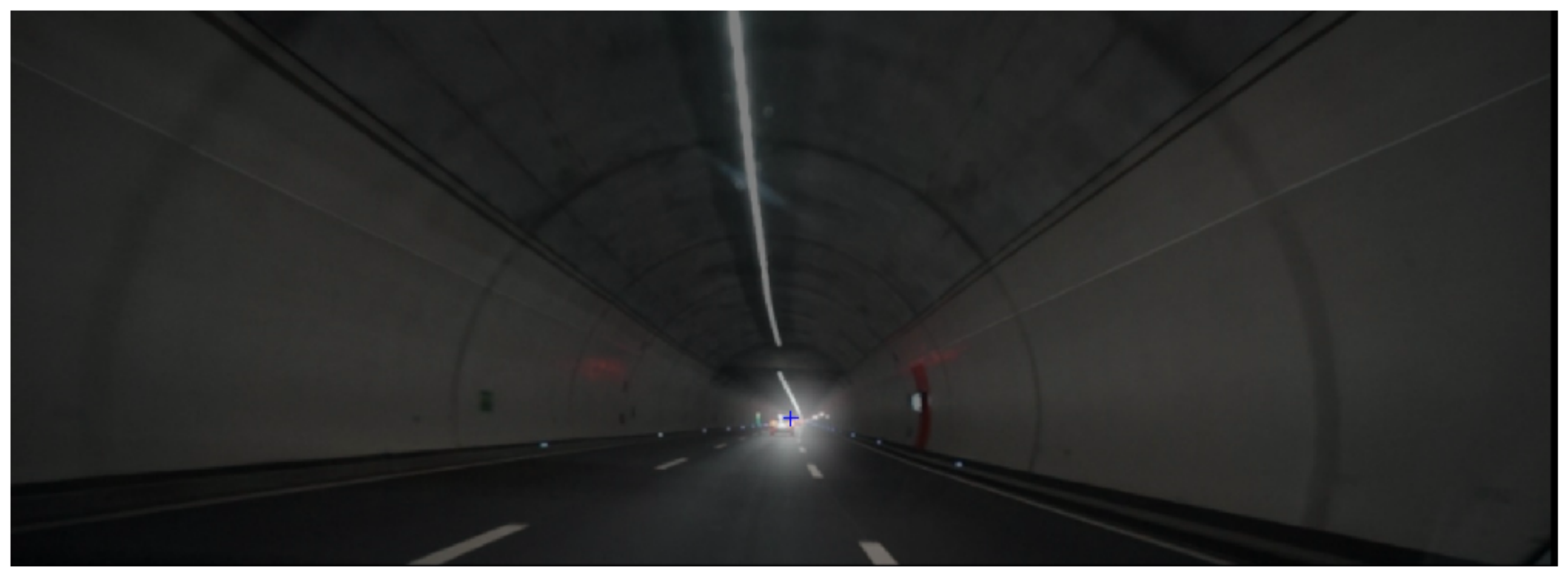}
  \end{subfigure}\\
\end{center}
   \caption{VP detection examples on CULane dataset with (top) and without (middle) lane markers and our test images (bottom). Probability distribution of the VP is visualized on the input image as a white heat-map and the detected VP as a blue plus mark. VP Detector has been trained with the CULane dataset with VP labels generated by 3-d curve fitting.}
 \label{figure:moments}
\end{figure*}

Figure \ref{fig:first_result} shows the evaluation result of the baseline ERFNet model trained with 3-dimensional curve fitting. For more than 95 \% of the test data, the NormDist error is less than 0.02. The percentage is significantly higher than that of the prior work \cite{liu2020heatmapbased} which is approximately 76\%, even if we consider the unlabeled test data (12\%). The mean NormDist values for the test-data subsets : i) the filtered test data subset where \(N_{int}\geq3\), \(\sigma_{vpy} < 10\) (\(N=26920\)) and ii) all the test data with at least two lane markings - \(N_{int}>0\) (\(N=30708\)) are 0.00629 and 0.00449 respectively. \(N_{int}\) stands for the number of intersection points and \(\sigma_{vpy}\) is the standard deviation of the intersection points along the y axis.
As is shown in Fig. \ref{figure:moments}, VP Detector is capable of predicting the VPs on our test images with diverse scenes, including a snowy scene (top right), a night and tunnel scene (middle left and right) and non-urban scenes (bottom). The result on the curved road (bottom right) shows that the detector has limitations in following the abrupt curve due to the lack of training data for such conditions.

\subsection{Comparison with Manual Annotations on VPG-DB-5ch dataset}
\begin{figure}[t]
\begin{center}
  \begin{subfigure}[b]{.325\linewidth}
    \centering
    \includegraphics[width=.999\linewidth]{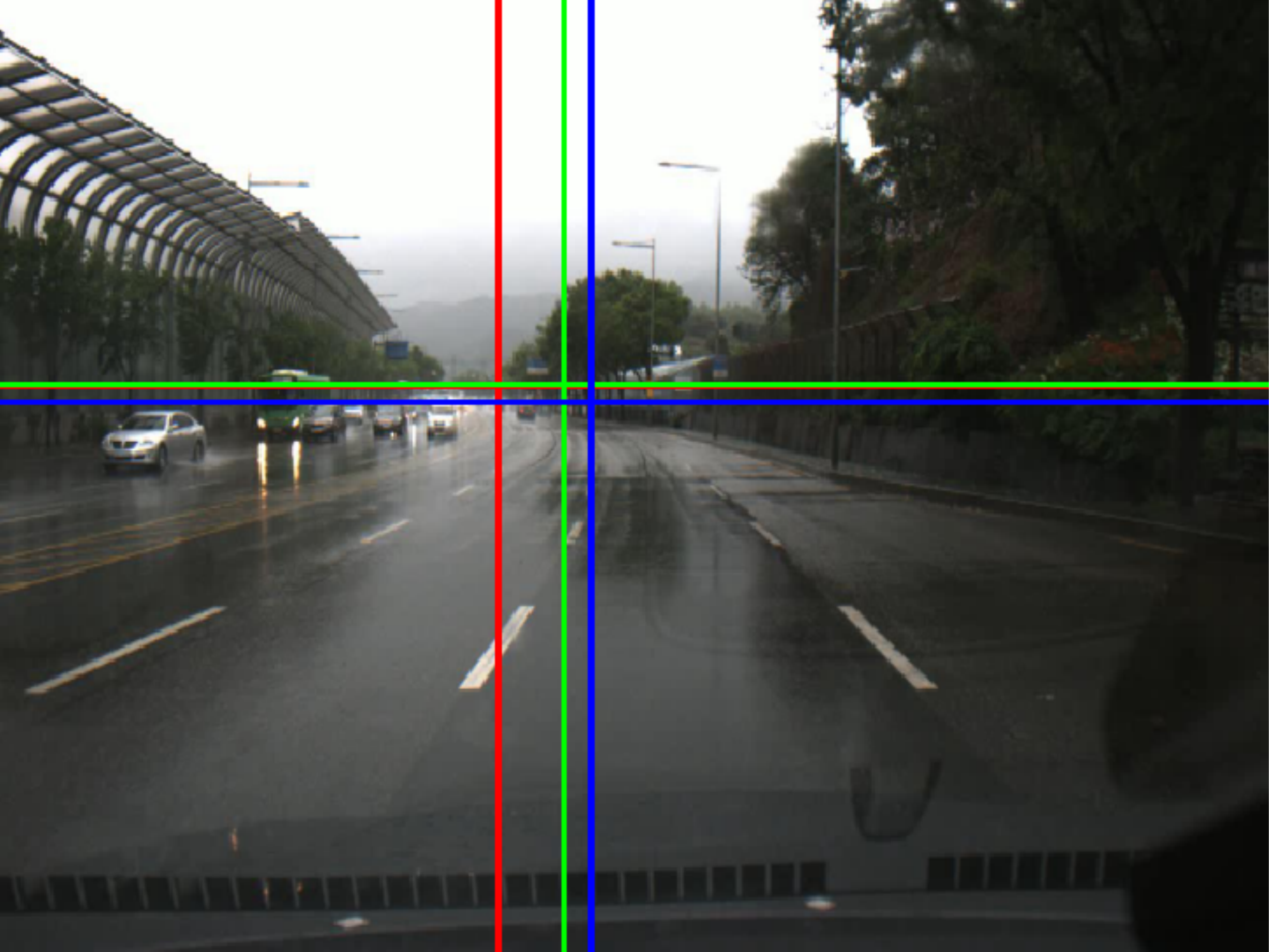}
    \caption{}
  \end{subfigure}
  \begin{subfigure}[b]{.325\linewidth}
    \centering
    \includegraphics[width=.999\linewidth]{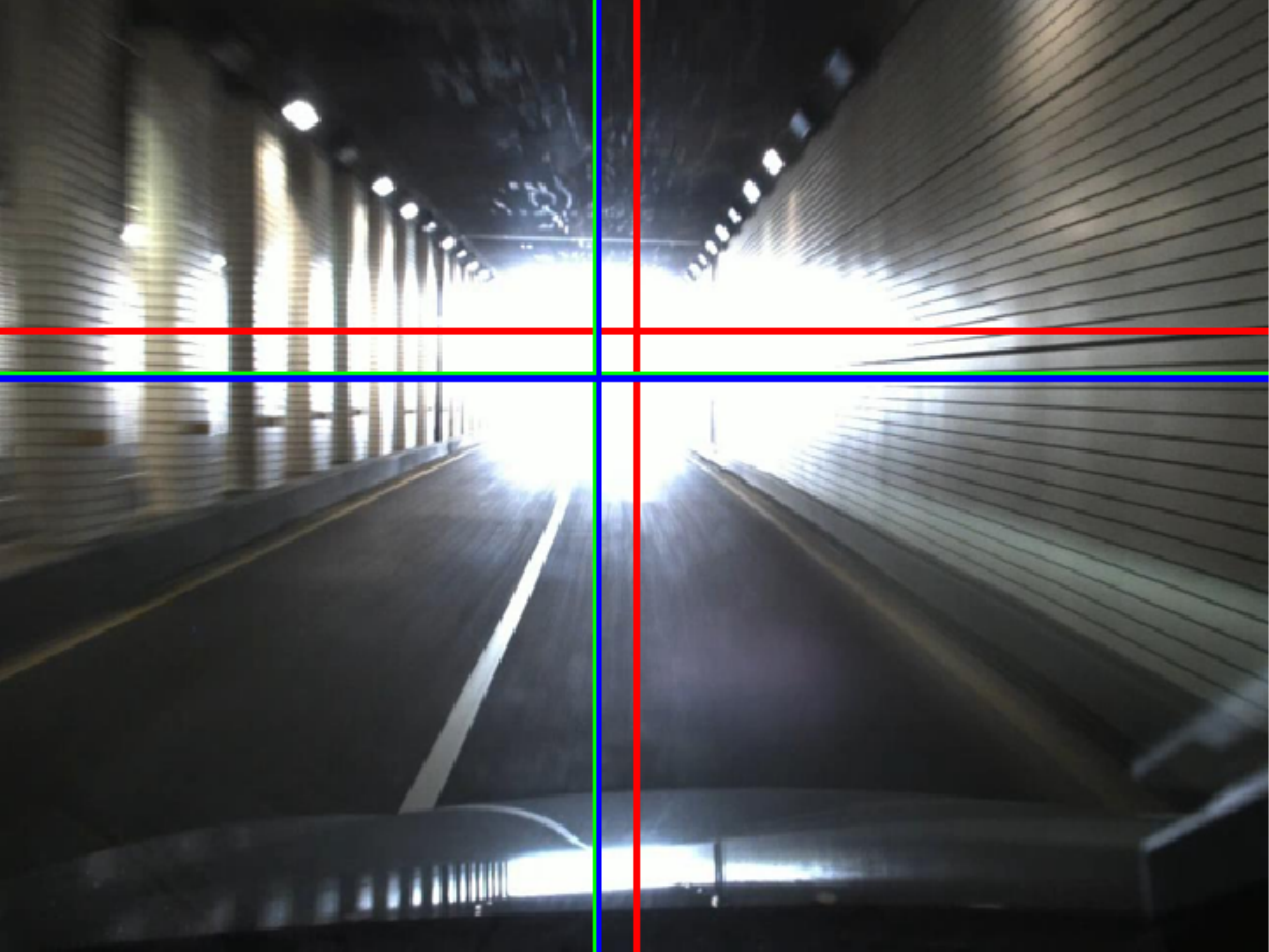}
    \caption{}
  \end{subfigure}
  \begin{subfigure}[b]{.325\linewidth}
    \centering
    \includegraphics[width=.999\linewidth]{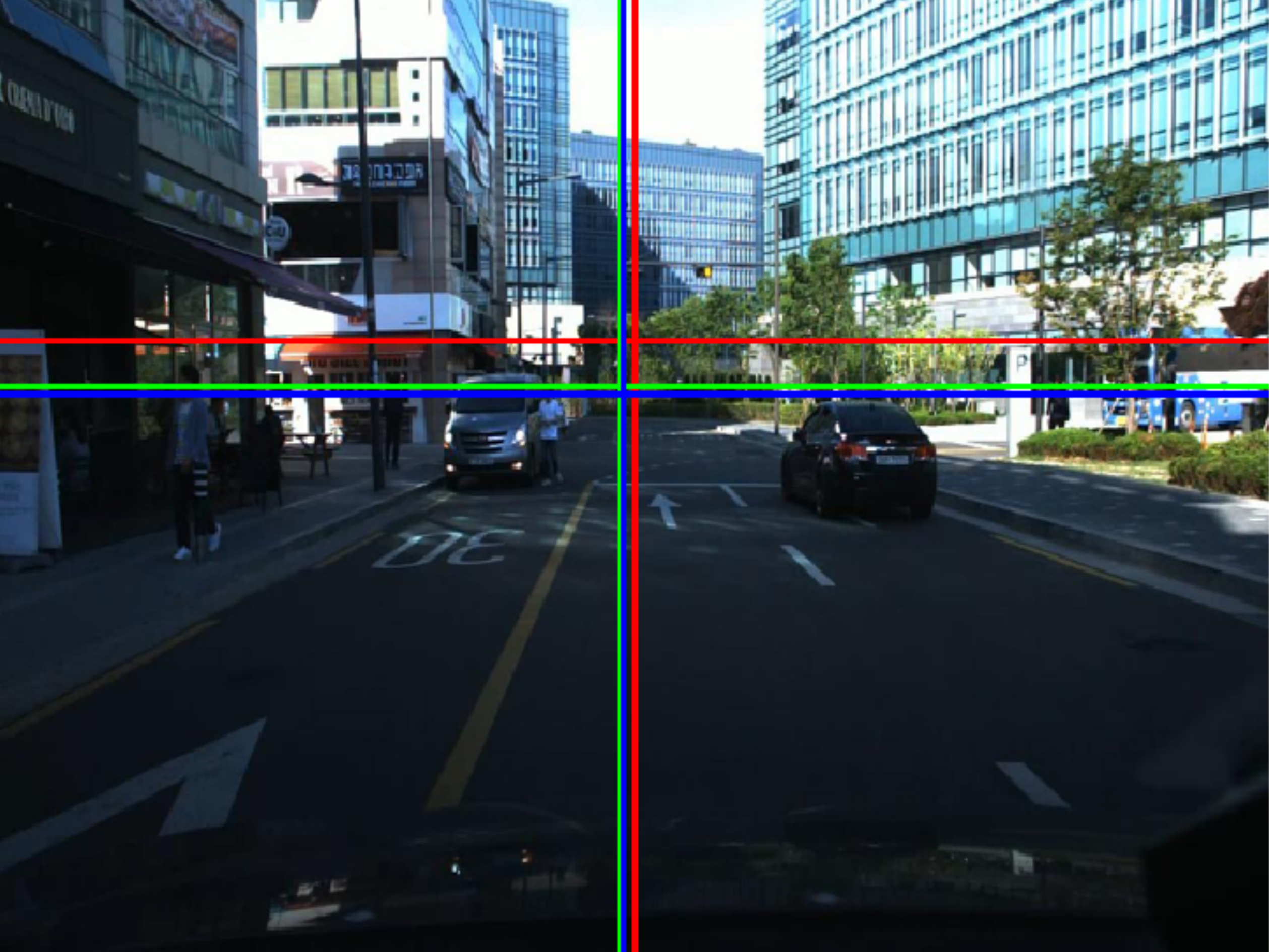}
    \caption{}
  \end{subfigure}
\end{center}
   \caption{Comparison of VP detection results (blue), automatic labels (green) and manual labels (red) on the VPG-DB-5ch dataset. The model has been trained with automatic labels. The manual labels are (a) on the left of and (b) (c) above the automatic labels.}
 \label{figure:vpg}
\end{figure}

We compare our automatic VP labeling results with manual annotations and also evaluate the generalization capability of the model on the VPGNet-DB-5ch dataset. \\
\textbf{Label Generation and Training.} We have trained ERFNet on two types of data subsets. For the subset (a), the data are filtered with the label generation results of \(N_{int}\geq3\) and \(\sigma_{vpy} < 10\). For the subset (b), the data which have valid manual annotations are selected. Automatic annotation labels are available only on (a). The input resolution is fixed to \((height, width) = (480, 640)\).\\
\textbf{Quantitative Evaluation.} 
The proportions of the frames where the NormDist is less than 0.02 (corresponding to 16 pixels) for all the train-test pairs are shown in Table \ref{table:vpg_auto_vs_manual}. Firstly, for all the cases VP accuracy is considerably higher than \cite{Lee_2017_ICCV} whose proportion of the true frames at the pixel distance of 16 is approximately 20\%. The VP Detector achieves significantly higher accuracy on the train-test pair with automatic annotation labels than the manual annotation cases. The result suggests that our automatic labeling is more stable than human labeling and mitigates annotation errors.\\
\textbf{Qualitative Evaluation.} 
Fig. \ref{figure:vpg} shows the visual examples of the VPGNet-DB dataset where the manual and the automatic labels differ. (a) : the manual label is on the left, where the curved road supposedly goes at the very far range, which results in ambiguity of VP labels along x-axis. For (b) the tunnel exit is saturated and the VP cannot be labeled correctly and for (c) the manual label is not consistent with lane markers, both of which result in VP errors along y-axis. These ambiguity and noise in manual annotations corroborate the quantitative evaluation results that only the train-test pair with automatic labels achieves low errors along both x and y axes. 
\iffalse
\begin{table}
\begin{center}
\begin{subtable}{0.7\textwidth}
\begin{tabular}{|c|c|c|c|}
\hline
 & \multicolumn{3}{c|}{Test}\\
 \cline{2-4}
 & Automatic& Manual &  Manual\\
Train & (a) N=863 & (a) N=863 & (b) N=3963 \\
\hline
Automatic & \textbf{0.9258} & 0.7358& 0.5132\\
(a) N=3793&  & &\\
\hline
Manual   & 0.7822 & 0.7254 & 0.5579\\
(a) N=3793  &  & &\\
\hline
Manual   & 0.7752 & 0.7879 & 0.6013\\
(b) N=15358  & & & \\
\hline
\end{tabular}
\end{subtable}
\end{center}
\caption{Evaluation results of the detectors trained with automatic and manual labels. Each model is evaluated on the test split with automatic and manual labels. Subset (a): the data are filtered with the label generation results of \(N_{int}\geq3\) and \(\sigma_{vpy} < 10\). Subset (b): the data which have valid manual annotations are selected.}
\label{table:vpg_auto_vs_manual}
\end{table}
\fi

\begin{table}
\begin{center}
\begin{subtable}{0.7\textwidth}
\begin{tabular}{|c|c|c|c|}
\hline
 & \multicolumn{3}{c|}{Test}\\
 \cline{2-4}
 & Automatic& Manual &  Manual\\
Train & (a) N=863 & (a) N=863 & (b) N=3963 \\
\hline
Automatic & \(x : \) \textbf{5.05} / & 11.39 / & 23.25 / \\
(a) N=3793& \(y : \) \textbf{4.11} & 6.45 & 12.23\\
 & (\textbf{92.6} \%) & (73.6 \%) & (51.3 \%) \\
\hline
Manual   & 9.48 / & 11.46 / & 23.07 /\\
(a) N=3793  & 5.76 & 6.71	& 12.32\\
 & (78.2\%) & (72.5 \%) & (55.8 \%) \\
\hline
Manual   & 10.04 / & 10.15 / & 18.64 /\\
(b) N=15358  & 5.95 &6.05 & 10.51 \\
 & (77.5 \%) & (78.8 \%) & (60.1 \%) \\
\hline
\end{tabular}
\end{subtable}
\end{center}
\caption{Evaluation results of the detectors trained with automatic and manual labels. MAEs and proportion of frames where \(NormDist < 0.02\) is shown in each cell. Each model is evaluated on the test split with automatic and manual labels. Subset (a): the data are filtered with the label generation results of \(N_{int}\geq3\) and \(\sigma_{vpy} < 10\). Subset (b): the data which have valid manual annotations are selected.}
\label{table:vpg_auto_vs_manual}
\end{table}

\subsection{Comparison with Lane Detection Result Fitting}
\begin{figure}[t]
\begin{center}
    \centering
    \includegraphics[width=.99\linewidth]{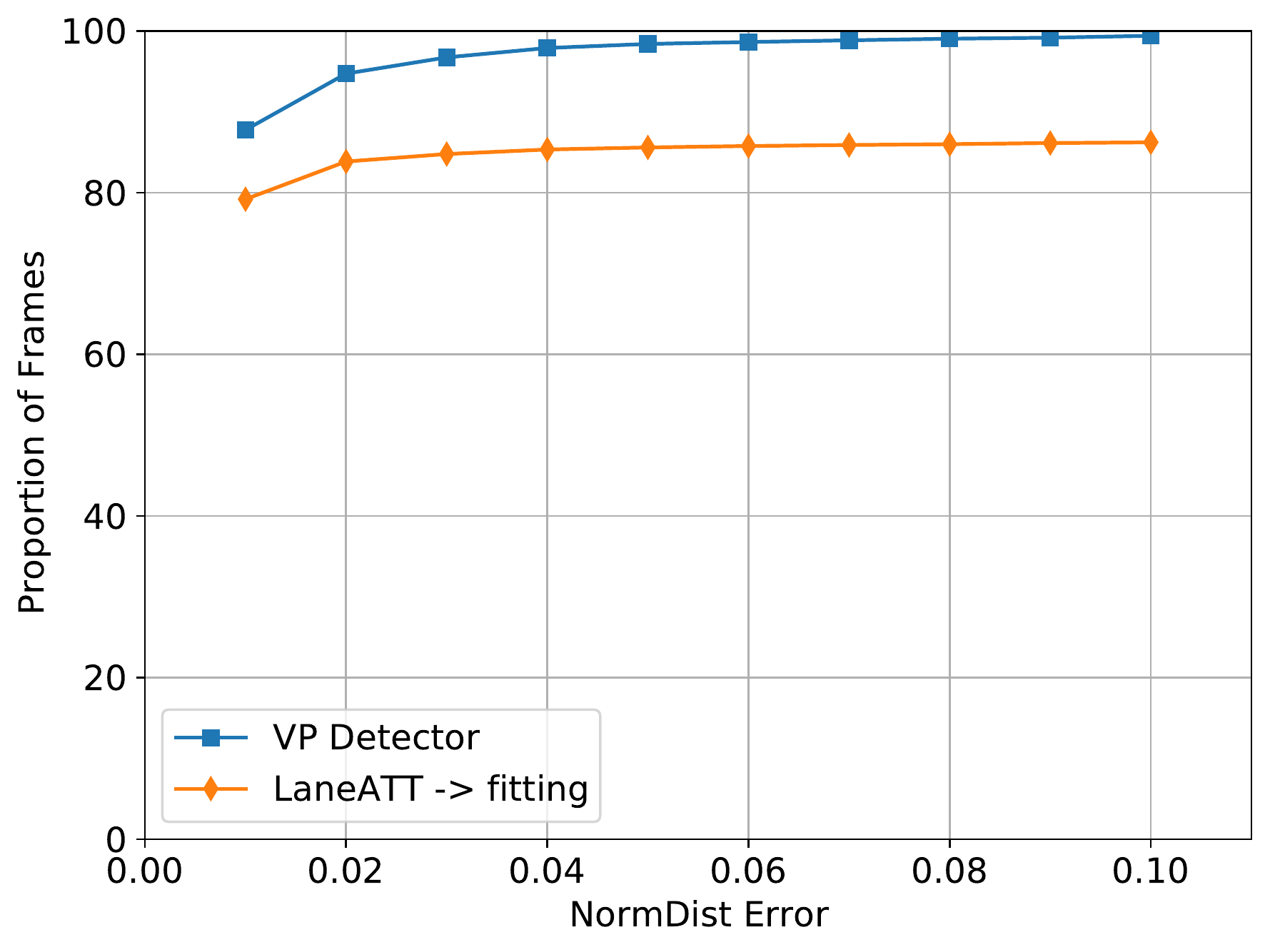}
\end{center}
   \caption{NormDist error comparison of VP detection accuracy between fitting of lane detection results and direct VP detection on the 30723 frames that have valid automatic labels.}
 \label{figure:laneatt}
\end{figure}

We validate the effectiveness of our VP Detector over the two-stage method where VP is calculated by fitting the lane detection results. As a lane detector, we adopt the off-the-shelf LaneATT detector \cite{tabelini2021cvpr} with the ResNet-122 backbone, which has 77\% of F1 score on the CULane dataset. The 3D curve fitting is applied on the lane detection results and the VPs are extracted. Fig. \ref{figure:laneatt} and Table \ref{table:laneatt} show the comparison of NormDist error between VP Detector and LaneATT result fitting. The cases in which the lane detector fails in detecting more than two lane markers are considered as \(NormDist = \infty\). VP Detector surpasses LaneATT result fitting by large margin mostly due to the lane detection failure cases. Even for the cases where there are sufficient lane detection results, VP Detector is more accurate (see Table \ref{table:laneatt} (a)). The results indicate that our direct VP detection overcomes the two-stage method by mitigating lane detection errors.

\begin{table}
\begin{center}
\begin{subtable}{0.7\textwidth}
\begin{tabular}{|c|c|c|c|c|}
\hline
 & \multicolumn{2}{c|}{NormDist $<$ 0.01} & \multicolumn{2}{c|}{NormDist $<$ 0.02}\\
  \cline{2-5}
Method & (a) & (b) & (a) & (b) \\
\hline
VP Detector & \textbf{0.9275} & 0.8777 & \textbf{0.9774} & 0.9474 \\
LaneATT fitting   & 0.9164 & 0.7919 & 0.9704 & 0.8385 \\
\hline
\end{tabular}
\end{subtable}
\end{center}
\caption{Comparison of VP detection accuracy between direct VP detection and lane detection result fitting on different subsets of CULane test data where (a) lane detector detects more than two lanes and (b) there are more than two annotated lane markers.}
\label{table:laneatt}
\end{table}

\subsection{Ablation Study}

\begin{table} 
\begin{center}
\begin{tabular}{|c|c|c|c|c|c|c|}
\hline
 &\multicolumn{3}{c|}{MAE (\(vp_{x}\))} & \multicolumn{3}{c|}{MAE (\(vp_{y}\))} \\
 \cline{2-7}
 &\multicolumn{6}{c|}{Test fitting}  \\
 \cline{2-7}
Train &1D & 1D- & 3D &1D & 1D & 3D  \\
fitting && close & & & close &  \\
\hline
1D & 7.72 & 9.05 &  10.49 & 4.14 & 5.51 &  4.36 \\
1D-close & 8.96 &  10.08 & 12.04 & 4.50 & 5.7 &  4.97\\
3D & 9.05 & 10.8 &  9.21 & 4.11 & 5.61 &  4.12\\
\hline
\end{tabular}
\end{center}
\caption{Mean average error (MAE) evaluation for the models trained with 1D, 1D-close, and 3D label settings. 
Each evaluation is carried out on the test data with each label fitting setting.
For all the models, input resolution is \((height=295, width=820)\) and MAE is calculated at the resolution scale of \((height=590, width=1640)\).} 
\label{table:accuracy_matrix}
\end{table}

\begin{table*}
\begin{center}
\begin{tabular}{|c|c|c|c|c|c|c|c|c|}
\hline
Input resolution  & Gaussian & Gaussian & Shift aug. & MACS & MAE  & MAE  &  \multicolumn{2}{c|}{\% of data where NormDist} \\
 (H, W)   & peak value & \(\sigma\) & probability & for test  & of \(vp_{x}\)&  of \(vp_{y}\) & $<$ 0.01 &  $<$ 0.02\\
\hline
(295, 820) & Fixed &  Fixed & 0.5 & 13.7 G &9.21 & 4.06& 87.8 \% & 94.8 \%\\
(295, 820) & Fixed &  Fixed & \textbf{0.0} & 13.7 G &9.21 & 4.06& 83.1 \% & 92.8 \%\\
(295, 820) & \textbf{Dynamic} &  Fixed & 0.5 & 13.7 G & 9.80 & 4.40& 87.1 \% & 94.2 \%\\
(295, 820) & \textbf{Dynamic} &  \textbf{Dynamic} & 0.5 & 13.7 G &10.95 & 4.46 & 86.6 \% &93.8 \%\\
\textbf{(160, 416)}  & Fixed &  Fixed & 0.5 & 3.74 G & 11.52 & 5.67 & 82.0 \% & 93.2 \%\\
\textbf{(80, 208)} & Fixed &  Fixed & 0.5 & 0.93 G &18.61& 9.29 & 62.3 \% & 85.3 \%\\
\textbf{(40, 104)} & Fixed &  Fixed & 0.5 & 0.23 G & 29.42 & 11.61 & 40.8 \% & 70.5 \%\\
\hline
\end{tabular}
\end{center}
\caption{Ablation study of VP Detector on the CULane dataset with 3d-fitting labels. The ERFNet model is trained with various conditions of shift augmentation probability, target Gaussian peak and \(\sigma\), and input resolution. The conditions that are modified from the baseline (at the top row) are highlighted. The results are compared w.r.t. number of multiply–accumulate operations (MACS), mean average error (MAE) at the resolution of \((height=590, width=1640)\) and the proportion of the frames where NormDist is less than 0.01 and 0.02.}
\label{table:ablation}
\end{table*}

%\begin{table}
%\begin{center}
%\begin{subtable}{0.7\textwidth}
%\begin{tabular}{|c|c|c|c|c|c|c|}
%\hline
%fitting  & Normal & Crowded & Night & Curve & Curve  \\
%degree & &&&& {\dag}\\
%\hline
%1d &99.4&96.6&95.1&85.1&93.9\\
%2d &99.5&96.3&94.3&78.8&89.6\\
%3d &99.0&96.1&92.6&70.8&87.6\\
%\hline
%\end{tabular}
%\end{subtable}
%\end{center}
%\caption{Evaluation results on test data subsets of different road conditions. The proportions of frames where NormDist error is less than 0.02 are shown. The Curve {\dag} stands for the result on the test data subset where \(N_{int}\geq3\) and \(\sigma_{vpy} < 10\).}
%\label{table:conditions}
%\end{table}

%  \begin{subfigure}[b]{.49\linewidth}
%   \centering
%    \includegraphics[width=.99\linewidth]{00281_Moment_trim_exp21}
%  \end{subfigure}
%  \begin{subfigure}[b]{.49\linewidth}
%    \centering
%    \includegraphics[width=.99\linewidth]{00281_Moment_1_trim_exp21}
%  \end{subfigure} \\
% \label{figure:moments}
%\end{figure*}
The accuracy of our VP Detector is evaluated on the CULane test set with different experiment settings.\\
\textbf{Curve Fitting Settings.}
We compare the models trained with labels generated by 1D, 2D, 3D and 1D (close range) curve fitting settings as shown in Table \ref{table:accuracy_matrix}. The models are evaluated on the test set with the four curve fitting settings respectively. For all the models, input resolution is \((height=295, width=820)\). The MAE results of \(vp_{y}\) do not differ among the fitting settings except for the 1D-close setting. We consider that the result is due to VP label fluctuation that is caused by less annotation points picked for fitting. As for \(vp_{x}\), the difference of fitting degrees at training and testing causes large MAE, which is because \(vp_{x}\) labeled by polynomials with different degrees do not coincide in the curve scenes (see Fig. \ref{fig:find_vp}).  \\
%\textbf{Road Conditions.}
%CULane dataset provides road condition labels for the test split. We evaluate VP Detector trained with labels generated by 1D, 2D and 3D fitting on each condition (Table \ref{table:ablation}). The detector is robust for the crowded and night conditions, however accuracy drop is observed for the curve frames with high-degree fitting, which is due to the automatic annotation %errors. 
%\\
\textbf{Target Probability Map.}
As is described in Sec. \ref{section:vpdetector}, we compare three target Gaussian distribution settings, where 1) \(\sigma\) is fixed and peak value \(A=1\), 2) \(\sigma\) is dynamic and \(A=1\), and 3) both \(\sigma\) and \(A\) are dynamic. For fixed \(\sigma\) we set \(\sigma\) as 16 pixels. For dynamic \(\sigma\) settings, we clip \(\sigma\) within the range from 6 to 16 pixels. As shown in Table \ref{table:ablation}, we observe no improvement from the baseline by introducing the dynamic target generation.
\\
\textbf{Shift Augmentation.}
The ablation study regarding the shift augmentation introduced in Section \ref{section:shiftaugmentation} is shown in Table \ref{table:ablation}. The lack of the augmentation degrades the performance of VP Detector. The training data of the CULane dataset has a limited number of camera settings, therefore without the augmentation the model is strongly overfit with respect to learning \(vp_{y}\).
\\
\textbf{Input Resolution.}
To evaluate the relationship between accuracy and computation cost, we have trained the ERFNet models with various input resolution settings: 1/4, 1/8 and 1/16 of the original resolution. The input image is padded horizontally and vertically with zeros to make its width and height divisible by 128, since the minimum feature map size for the minimum input resolution setting is 1/128 of the original resolution. For the resolution settings above, \(\sigma\) is fixed to 8, 4 and 2 respectively. The evaluation results in Fig. \ref{table:ablation} shows that the detector maintains comparable accuracy for the (160, 416) resolution and starts to degrade at the smaller resolutions.
\\
\textbf{Confidence Thresholding.}
From each frame, the VP is detected accompanied by a confidence value, which is the peak value of the probability map. As is the case in detection tasks, the higher the confidence is, the more accurate the VP location is expected to be. We evaluate the NormDist Error and number of available frames with confidence threshold of 0.0 and 0.99. The NormDist error and number of  available frames with confidence greater than 0.99 are 0.0021 (62\% drop) and 9,860 (68\% drop) respectively, compared with no-thresholding results. \\
\subsection{Horizon Line Estimation}
Lastly, we demonstrate the VP detector is capable of obtaining the horizon line when applied on a video. The CULane test set consists of three camera attachment settings. We conduct inference on the test set utilizing the VP detector trained with 3d-fitting labels. The confidence values - the peak value of the probability map - from the predictions are accumulated at the peak positions and the 1-d line fitting is conducted on the column-wise maximum positions. The calculated angles between the fitting results and the x-axis are (a): $1.49$ and (b): $1.00$ degrees respectively. 
 
\begin{figure}
\begin{center}
\begin{subfigure}[b]{.49\linewidth}
    \centering
   \includegraphics[width=0.99\linewidth]{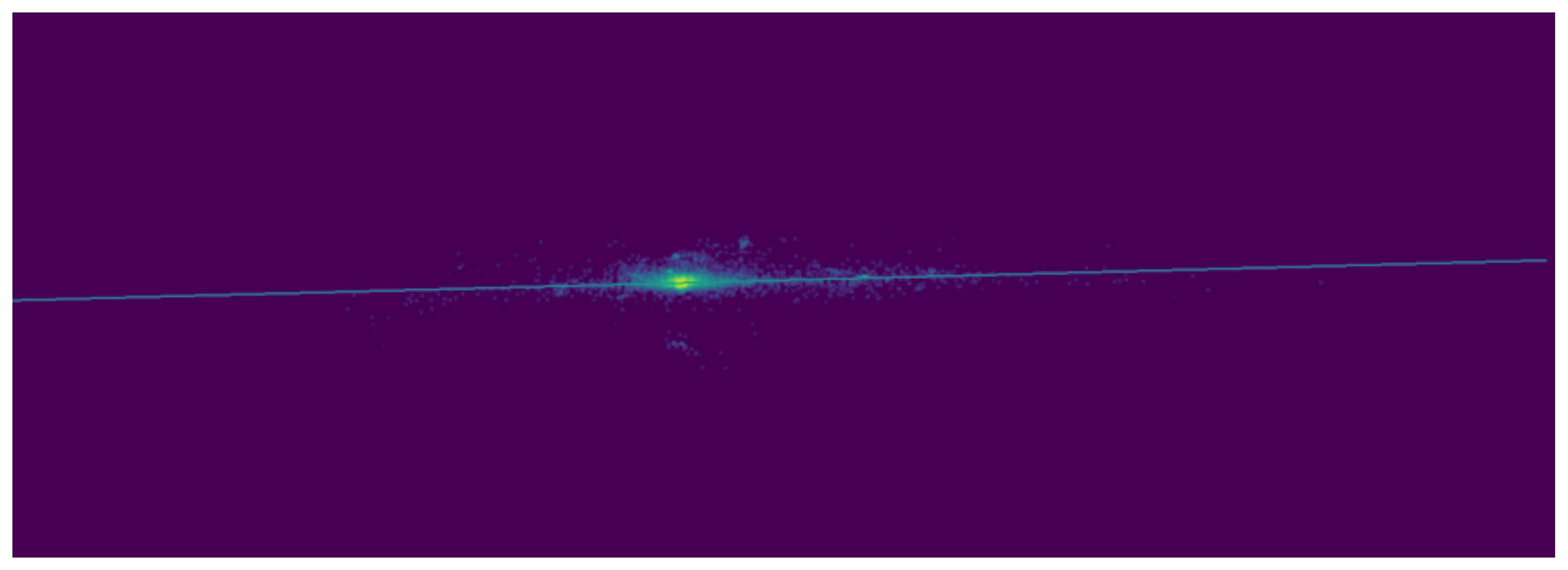}
   \caption{driver\_100\_30frame setting}
\end{subfigure}
\begin{subfigure}[b]{.49\linewidth}
   \centering
   \includegraphics[width=0.99\linewidth]{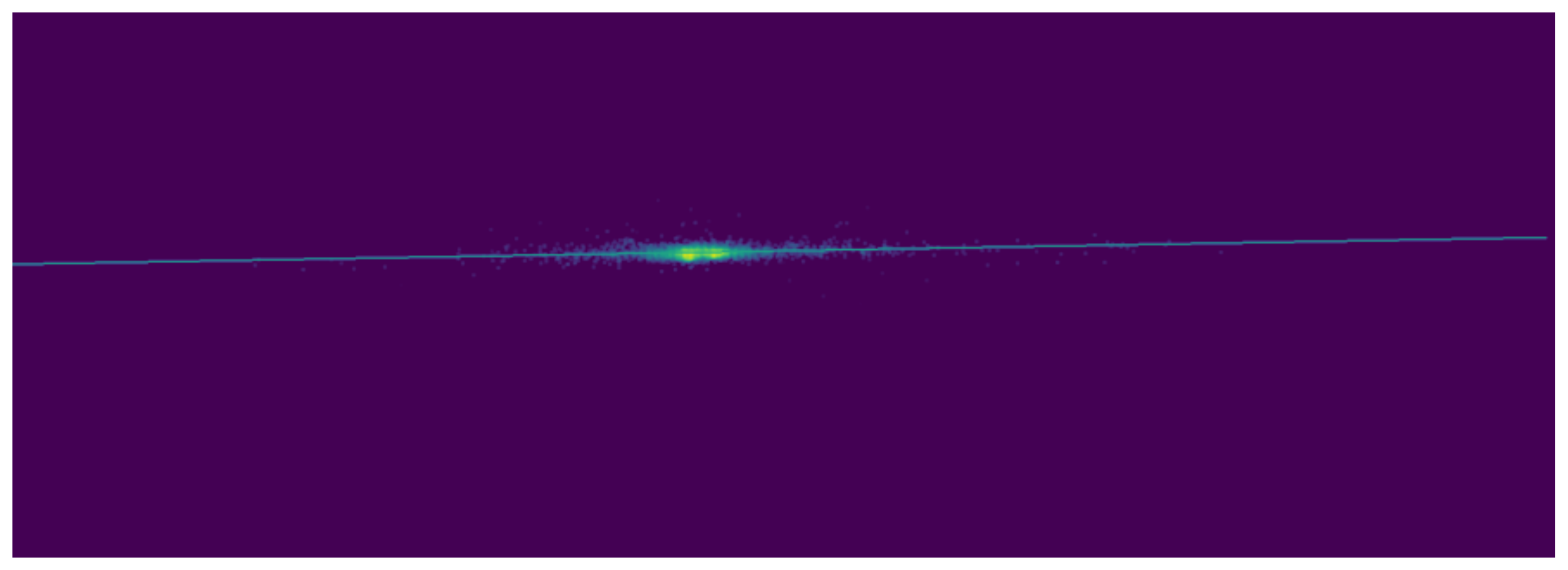}
   \caption{driver\_193\_90frame setting}
\end{subfigure}
\end{center}
  \caption{Horizon line estimation for two camera settings in the CULane test set. The peak value of the probability map for each frame is accumulated at each peak position.}
\label{fig:horizonline}
\end{figure}

%%%
%\begin{figure}[t]
%\begin{center}
%   \includegraphics[width=0.99\linewidth]{vpg_results.png}
%\end{center}
%  \caption{Evaluation results on VPG-DB-5ch dataset. Cumulative NormDist error on different \textbf{test data} subsets are shown - blue and orange: \(N_{int}\geq3\), \(\sigma_{vpy} < 10\) (\(N=4810\)) green: \(N_{int}>0\) (\(N=19321\)). Blue plot represents the test on the automatically extracted labels and the others manual labels.}
%\label{fig:vpg_results}
%\end{figure}
%%%

\section{Conclusion}

   We have proposed an end-to-end monocular vanishing point (VP) Detector trained with automatically labeled VPs. By calculating geometric intersections from the lane marker annotations, we are able to mitigate the VP annotation error and realize stable VP detection via heatmap estimation. We have demonstrated that our detector surpasses the methods based on lane detection and manual annotation in accuracy. We believe our method paves the way for accurate online camera calibration to realize stable driving scene recognition.
\clearpage

{\small
\bibliographystyle{ieee_fullname}
\bibliography{vpdetector}
}

\end{document}